%% file: iclr2027_conference.tex
\documentclass{article} 
\usepackage{iclr2027_conference,times}

\input{math_commands.tex}

\usepackage{hyperref}
\usepackage{url}
\usepackage{booktabs}   
\usepackage{multirow}   
\usepackage{graphicx}   
\usepackage{amssymb}    
\usepackage{natbib} 
\usepackage{hyperref}       
\usepackage{url}            
\usepackage{booktabs}       
\usepackage{amsfonts}       
\usepackage{nicefrac}       
\usepackage{microtype}      
\usepackage{multirow}
\usepackage{booktabs}
\usepackage{graphicx}
\usepackage{amsmath}
\usepackage{wrapfig}
\usepackage{subcaption}
\usepackage[dvipsnames]{xcolor}
\usepackage{enumitem}
\usepackage{algorithm,algorithmicx,algcompatible}
\usepackage{array}
\usepackage{algpseudocode}
\usepackage{float}
\usepackage{placeins}
\usepackage[normalem]{ulem}
\hypersetup{hidelinks}

\newcommand{\eg}{\textit{e.g.,~}}

\title{Correcting WHERE, Preserving HOW: Compositional Generalization for Vision-Language-Action Models via Referential Guidance}

\author{%
\textbf{Yanyan Zhang \quad
Disheng Liu \quad
Xinpeng Li \quad
Chaoda Song \quad
Mohsen Hariri}\\
\textbf{Debargha Ganguly \quad
Wang Yang \quad
Kai Ye \quad
Bryce Grant \quad
Vipin Chaudhary\textsuperscript{$\dagger$} \quad
Yu Yin\textsuperscript{$\dagger$}}\\[1mm]
Case Western Reserve University \\
Cleveland, OH, USA \\
\texttt{yxz3106@cwru.edu} \\
}

\iclrfinalcopy 
\begin{document}
\raggedbottom

\maketitle
\lhead{Correcting WHERE, Preserving HOW}
\begingroup
\renewcommand{\thefootnote}{\fnsymbol{footnote}}
\footnotetext[2]{Co-corresponding authors.}
\endgroup

\begin{abstract}

While Vision-Language-Action (VLA) models enable flexible action generation, their generalization across diverse environmental elements, including manipulated objects, destinations, and backgrounds, is limited by the lack of diversity in robotic training data. Trained end-to-end on such data, VLAs tend to exploit visual shortcuts, associating actions with task-irrelevant visual features rather than the intended task semantics. These shortcuts block recomposition of elements already seen by the policy, that is, \textit{compositional generalization}. Existing approaches mitigate such entanglement through task-relevant perception or targeted data diversification, but offer no explicit mechanism for unseen recomposition and require backbone-specific modifications with retraining. We observe that under such recomposition, VLAs often fail at global grounding while retaining local manipulation skills that recover near the correct target in familiar configurations. Therefore, we propose \textbf{Referential Guidance (ReGuide)}, a training-free wrapper that, given object poses from a grounding module, combines semantic and geometric rebinding to guide the end-effector into demonstration-supported configurations of the instructed referent, where the frozen policy can resume execution. Experiments in simulation across multiple VLA backbones as well as on a real robot show that ReGuide improves success rates under compositional shifts by up to 56.8 and 75.0 percentage points, respectively, while preserving standard-task performance.

\end{abstract}

\section{Introduction}

Vision-Language-Action (VLA) models draw on the broad semantic knowledge from pretrained vision-language backbones and map visual observations and language instructions directly to robot actions~\citep{OpenVLA,RT2}. However, due to the high cost of robotic data collection, objects, destinations, and backgrounds in robotic datasets tend to appear in fixed combinations~\citep{Shortcut,saxena2025what,zhou2026libpro}. Trained end-to-end on such data with direct low-level output, VLAs receive supervision only on the final action, where the contributions of object, skill, and environment are collapsed into raw action values~\citep{hancock2026actions,pi05}. Which part of an action belongs to the interaction with the \textit{target object/destination}, which to the \textit{skill} itself, and which to the constraints of the \textit{environment} is hard to disentangle, since these factors rarely vary independently in training datasets~\citep{gao2024}. As a result, the policy often fails when a test episode breaks a training combination (\eg manipulating an object in a different scene, or pairing a target object with a new destination), even though it has seen every element before~\citep{Shortcut}. It binds to the wrong referent, disregards the instruction, and blindly replays a training trajectory anchored on the dominant visual cues~\citep{NEURIPS2019_94701864,AC-VLA,hou2026langgap}. The compositional generalization that VLAs should in principle possess is thus limited by robotic data and end-to-end training~\citep{chen2025robohiman}. 

\begin{figure}[t]
    \centering
    \includegraphics[width=\linewidth]{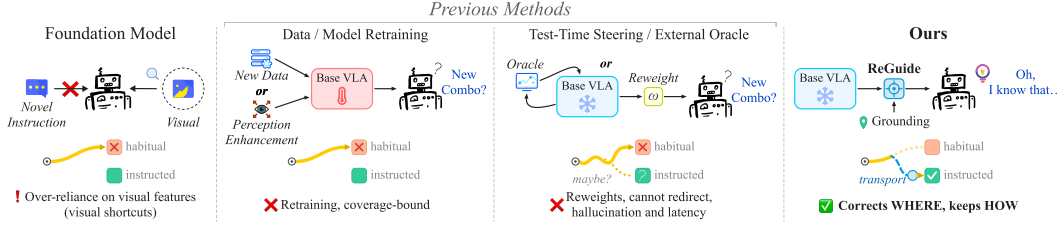}
    \caption{\textbf{Approaches to compositional failures.} Foundation VLAs over-rely on visual features. Previous methods require coverage-bound retraining or use test-time steering and external oracles, while ReGuide corrects the referent and preserves the frozen policy’s manipulation competence. The second and third panels from the left correspond to strands 1--2 and strand 3 respectively.}
    \label{fig:teaser}
\end{figure}

Recent work on this limitation falls into three strands, summarized in Fig.~\ref{fig:teaser}. The first decorrelates the factors in the data, collecting or synthesizing demonstrations in which objects, goals, and scenes are re-paired~\citep{vo2025clutter,gao2024,Shortcut,Unleashing,AC-VLA}. Such methods buy composition with coverage, so the cost grows with the number of factors, and pairings outside the collected coverage still fail~\citep{xie2024decomposing}. The second changes how the policy grounds the referent, by injecting explicit target cues~\citep{li2025controlvla}, or by separating declarative from procedural knowledge in a new architecture~\citep{Decoupling,lian2026langforce,huang2025otter}. These methods retrain the policy to incorporate grounding cues, often with backbone-specific modifications~\citep{vo2025clutter}. Generalization to unseen pairings still depends on what the retrained mapping covers~\citep{RoboGround}. A third line of research keeps the policy frozen and steers it at inference. Some works amplify the language-dependent component of the policy's own action distribution, which only reweights what the policy already proposes and cannot carry the end-effector to a referent it never considered~\citep{zhan-etal-2026-stable,Confined,fang2026vision}. Others keep an external oracle in the loop, such as learned action scorers or language-model planners staging the robot~\citep{nakamoto2024steering,zhang2026harness,du2025dyna}. Every correction then depends on that stack, which adds latency and, for generative planners, the risk of infeasible subgoals~\citep{Irpan2022,li2026large}. Across all three strands, the failure is treated as missing knowledge to be rebuilt or as a distribution to be reweighted or overridden, without taking a closer examination: \textit{which part of the behavior actually fails, and which part still works? Can its competence and flexibility stay in charge with minimal agentic assistance?}

Our observation is that failures usually stem from incorrect global grounding rather than a loss of local manipulation skills. The policy commits to the wrong referent, yet once its end-effector reaches a trained configuration, it can resume local interaction with the object. The policy often errs in \textit{where} it acts while retaining the skills needed for \textit{how} it interacts. We therefore propose \textbf{Referential Guidance (ReGuide)}, a training-free wrapper which corrects a wrong referent commitment by rebinding rather than retraining or reweighting the policy. Specifically, the rebinding consists of two components. The semantic component determines which referent to act on, while the geometric component determines the local pose from which the policy can take over. On the \textit{semantic side}, it resolves the instructed referent via a lightweight grounding module and uses the policy's action predictions as a signal for detecting referent mismatches without an external scorer. On the \textit{geometric side}, it transports the end-effector along a bounded flow into a demonstration-supported pre-contact set of the correct referent. The frozen policy then resumes local interaction from the true observation, with subsequent actions constrained by the demonstrated motion envelope. ReGuide derives its reference configurations and motion bounds from the policy's training demonstrations. Our contributions are as follows:

\begin{itemize}[nosep,leftmargin=1.3em]
    \item We find that VLAs retain local manipulation skills under compositional shifts, which can be reused by guiding the end-effector into demonstration-supported configurations of correct referents.
    
    \item We propose \textbf{Referential Guidance (ReGuide)}, a training-free, backbone-agnostic wrapper that reuses local interaction skills through \textit{semantic} and \textit{geometric rebinding}.
    
    \item ReGuide raises compositional success by up to 56.8 percentage points in simulation and 75.0 points on a real robot, while preserving standard-task performance.
\end{itemize}

\section{Related Work}

\subsection{Vision-Language-Action Models}

VLA models fine-tune a pretrained VLM to map images, proprioception, and an instruction directly to robot actions~\citep{X-Embodiment}. Early designs emit actions as discrete tokens through autoregressive decoding~\citep{OpenVLA,RT2,pertsch2025fast}, while recent generalist policies attach a diffusion or flow-matching action expert that outputs continuous action chunks~\citep{black2026pi0,zheng2026xvla,nvidia2025gr00t,shukor2025smolvla}, and fine-tuning methods further improve inference speed and task success~\citep{kim2025fine}. Their generalization is inherited from the VLM backbone~\citep{RT2}, yet fine-tuning on robot data that is far less diverse than the backbone's pretraining corpus weakens the grounding it started with~\citep{huang2025otter}. This has motivated frozen encoders and knowledge insulation during training~\citep{driess2025knowledge}, and test-time sampling with verifiers that improve robustness without touching the weights~\citep{kwok2025robomonkey,nakamoto2024steering}. ReGuide leaves the model, its training, and its decoding untouched and wraps any of these policies as a frozen executor.

\subsection{Compositional Generalization in Robot Manipulation}

Compositional generalization in manipulation can mean composing skills into sequences~\citep{chen2025robohiman,yang2026lilo} or, as studied here, recombining task elements~\citep{gao2024,Confined}. We organize prior works by whether they act during training or at test time.

\noindent\textbf{Training-time methods.}
Data-side work attributes shortcuts to limited factor combinations rather than insufficient data~\citep{Shortcut}, and answers with protocols covering factor pairs~\citep{gao2024} or synthesis pipelines that assemble recombined trajectories by LLM decomposition with wrist masking~\citep{AC-VLA} or by steered rollouts filtered in simulation~\citep{Unleashing}. Model-side work identifies the referent through zero-initialized cross-attention on object masks~\citep{li2025controlvla}, a lifted 3D point modulating the action head~\citep{Direct_Action-Head}, or object-centric tokens~\citep{bendikas2025focusing}, or reshapes attention through relevance supervision~\citep{GuidedVLA,sun2026artificial}, separate where and what streams~\citep{Decoupling}, and insulated or mutual-information objectives~\citep{driess2025knowledge,lian2026langforce}. 

\noindent\textbf{Inference-time steering.}
Inference-time methods differ in the signal they steer with.
Language-side steering interpolates text latents between tasks~\citep{Confined}, contrasts the policy against a language-dropped branch~\citep{fang2026vision}, or redistributes attention to instruction tokens~\citep{zhang2026restoring}, each always on with hand-set scales. Oracle-side steering re-ranks samples with a value or verifier~\citep{nakamoto2024steering,kwok2025robomonkey,wu2026inference}, guides denoising with a dynamics or reward model~\citep{du2025dyna,liu2026vls}, or stages the robot before contact, after a stall through an affordance field combining a planner model, a segmenter, depth, and a motion planner~\citep{xu2026affordance}, or through a memory-guided agent~\mbox{\citep{zhang2026harness}}. 

ReGuide instead needs no retraining, no external planner or scorer, and no guidance-weight tuning, directly reusing local manipulation skills across multiple VLA backbones.

\section{Methodology}
\label{sec:method}

\subsection{Problem Formulation}
\label{sec:pf}

\textbf{Setting.}
A frozen VLA policy $\pi_\theta$ maps an observation $o_t$ and an instruction $\ell$ to an action chunk $A_t$, executing a prefix before replanning. Each action $a=(u,w,g)$ contains translation, rotation-vector and gripper commands for the end-effector pose $x_t=(p_t,R_t)$. The instruction specifies an object and a destination, inducing stages such as grasping and placing. At stage $k$, the instructed referent is $e_k^\star$, and the moving point $q_t$ is the end-effector position before the object is held and the object position afterwards. A grounding module identifies the instructed referent and supplies entity poses and axis-aligned bounding boxes.

\begin{figure}[t]
    \centering
    \includegraphics[width=0.9\linewidth]{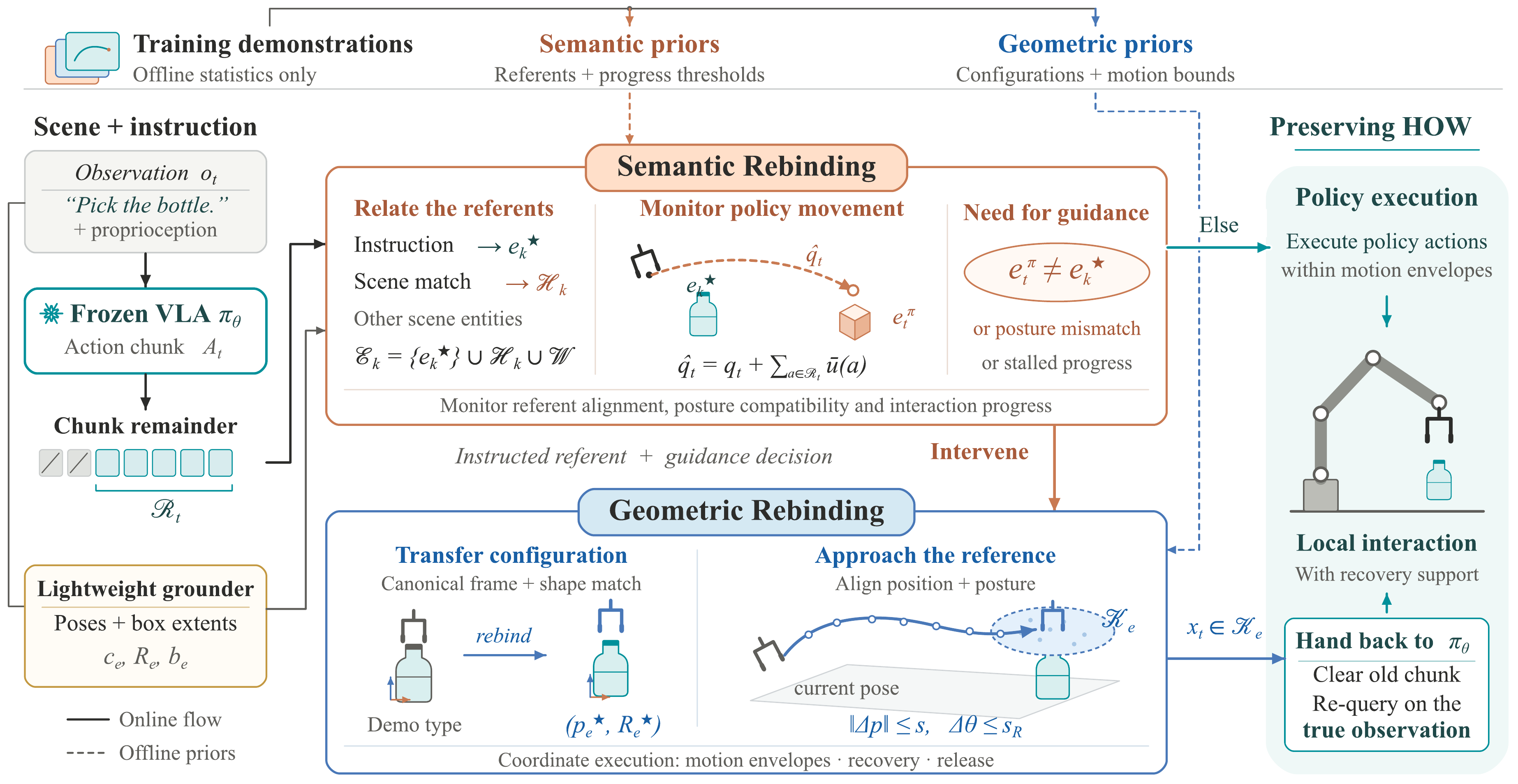}
    \caption{\textbf{Overview of ReGuide.} \textit{Semantic rebinding} monitors referent alignment and interaction progress. \textit{Geometric rebinding} transfers demonstration-derived configurations and coordinates guidance, frozen-policy execution and recovery.}
    \label{fig:framework}
\end{figure}

\textbf{Restoring the conditions for local interaction.}
A successful stage requires reaching the instructed referent and executing a compatible interaction. For stage $k$, let $B_k$ denote entry into the correct referent's approach region (Sec.~\ref{sec:semantic}) and $S_k$ denote stage success. Since success requires entry, with $\Pr$ denoting probability,
\begin{equation}
\Pr(S_k)=\underbrace{\Pr(B_k)}_{\beta_k\;\text{(binding)}}\;\underbrace{\Pr(S_k\mid B_k)}_{\gamma_k\;\text{(competence)}},
\label{eq:factor}
\end{equation}
where binding $\beta_k$ captures reaching the correct interaction and competence $\gamma_k$ captures completing it. The design objective is to restore access to the interaction configurations in which the policy's local skills remain usable. This requires both identifying the correct referent and approaching it in a suitable position and posture. Our framework addresses these requirements through two components (Fig.~\ref{fig:framework}). \textit{Semantic rebinding} relates the policy's movement to the instructed referent and determines when guidance is needed. \textit{Geometric rebinding} transfers a local interaction configuration to that referent and coordinates transport, hand-back and subsequent execution. For example, destination reassignment changes where the held object should go, while the demonstrated release configuration specifies how it should arrive. Appendix~\ref{app:binding_analysis} gives the corresponding intervention model.

\textbf{Demonstrations as references.}
The demonstrations $\mathcal{D}$ that trained $\pi_\theta$ supply target configurations, motion bounds and progress thresholds. For a scalar quantity $\xi$, $Q_\alpha[\xi]$ denotes its empirical quantile within a referent class. Targets use $Q_{.5}$, upper and lower bounds use $Q_{.95}$ and $Q_{.05}$, and commitment thresholds use $Q_{.25}$. These levels and the execution constants are shared across tasks (Appendix~\ref{app:demo_stats}).

\subsection{Semantic Rebinding}
\label{sec:semantic}

\textbf{Relating intended and competing referents.}
The instructed referent identifies the desired interaction. Alternatives reveal whether the policy follows a familiar but incorrect association. We retrieve the training scene nearest to the first frame under a frozen image encoder and collect the objects and destinations moved by its tasks into a habitual set $\mathcal{H}_k$. The candidates are $\mathcal{E}_k=\{e_k^\star\}\cup\mathcal{H}_k\cup\mathcal{W}$, where $\mathcal{W}$ contains the remaining movable entities. Each candidate inherits demonstration statistics from the training type matched by geometry rather than name.

\textbf{Monitoring the policy's referent.}
The unexecuted chunk remainder $\mathcal{R}_t$ predicts where the policy is moving. Translation and rotation commands are converted to expected increments by $\bar u=\kappa_p\sigma_p u$ and $\bar w=\kappa_R\sigma_R w$, using controller scales $\sigma$ and tracking gains $\kappa$ fitted on demonstrations. The predicted endpoint and posture are
\begin{equation}
\hat q_t=q_t+\sum_{a\in\mathcal{R}_t}\bar u(a),\qquad
\hat R_t=\exp\!\Big(\Big[\sum_{a\in\mathcal{R}_t}\bar w(a)\Big]_\times\Big)R_t .
\label{eq:endpoint}
\end{equation}
Commitment is assessed relative to each referent's approach geometry. For referent position $c_e$ and box half-extents $b_e$, the componentwise offset $\delta_e(q)=\max(|q-c_e|-b_e,0)$ is zero along axes where $q$ lies inside the box. The entry region has at most one nonzero component. Demonstration entry is the first frame before engagement in which the end-effector lies in this region. The commitment radius $\rho_e=Q_{.25}[\|\delta_e(p_t)\|]$ is measured at this event. A commitment requires the predicted endpoint to enter along two axes and approach along the remaining axis within this radius,
\begin{equation}
\chi_t(e)=\mathbb{1}\!\left[
\|\delta_e(\hat q_t)\|_0\le 1,\quad
\delta_e^{\,i}(\hat q_t)<\rho_e,\quad
\delta_e^{\,i}(\hat q_t)<\delta_e^{\,i}(q_t)
\right],
\label{eq:commit}
\end{equation}
where $i$ indexes the largest endpoint offset and components below the position resolution $\varepsilon$ count as zero. Among qualifying candidates, the box nearest to the predicted endpoint defines the policy's referent $e_t^\pi$. A mismatch triggers guidance when it differs from $e_k^\star$, the moving point remains at least $\rho_{e_k^\star}$ from the instructed referent's box, and the gripper has not contacted it. This provides an anticipatory mismatch signal without an external scorer. During placement, the comparison uses horizontal object--destination distance and the demonstrated carry radius, requiring  consecutive verdicts unless the object already dwells within a habitual radius. Grasp-stability and height-envelope gates coordinate this decision with the carrying state (Appendix~\ref{app:semantic}).

\textbf{Monitoring progress toward interaction.}
A policy can select no competing referent yet fail to approach a usable configuration. ReGuide monitors posture compatibility and progress toward the demonstrated grasp configuration. Before descent beside the object, guidance is triggered if the predicted posture error exceeds its demonstrated tolerance and improves over the current posture by no more than one demonstrated rotation step. Failure to bind is declared after $N$ steps without one demonstrated step of progress, with $N$ exceeding the longest such demonstration interval by one. An object with no matching training type receives guidance at the first decision using geometric transfer.

\subsection{Geometric Rebinding}
\label{sec:geometric}

\textbf{Transferring an interaction configuration.}
Given the selected referent, geometric rebinding supplies the position and posture from which to interact with it. Demonstrations are grouped by object type and destination class. For an object of orientation $R_e$, poses at engagement are expressed in its gravity-aligned canonical frame $R_eC_e$, whose first axis follows the wider horizontal extent and whose third is vertical. Let $t_e$ locate the top-face centre in this frame, $\bar o$ be the per-axis median grasp offset from it, and $\bar R$ the rotation medoid. Combining these with the median entry altitude $\bar z$ above the object top $z_e^{\mathrm{top}}$ gives
\begin{equation}
\begin{aligned}
p_{e,xy}^\star&=c_{e,xy}+\big[R_eC_e(\bar o+t_e)\big]_{xy},\qquad
p_{e,z}^\star=z_e^{\mathrm{top}}+\bar z,\\
R_e^\star&=R_eC_e\bar R.
\end{aligned}
\label{eq:target}
\end{equation}
An unseen object inherits statistics from the nearest training type under Euclidean distance between log descriptors of height, two widths and top fill. The canonical frame transfers the reference to its geometry (Appendix~\ref{app:geo}). 

Hand-back requires proximity in both position and posture. The pre-contact set is
\begin{equation}
\mathcal{K}_e=\Big\{(p,R)\ \Big|\ \|p-p_e^\star\|\le s,\ \ \|\Pi_\perp(p-p_e^\star)\|\le r_e,\ \ \vartheta(R_e^\star,R)\le\tau_R\Big\}.
\label{eq:preset}
\end{equation}
Here $s$ is the $Q_{.95}$ translation step. The lateral tolerance $r_e$ is the smaller of the $Q_{.95}$ grasp scatter along the closing axis and the object half-extent along it. The posture tolerance $\tau_R$ is the $Q_{.95}$ rotation residual at entry. The projection $\Pi_\perp$ removes the approach axis, and $\vartheta$ measures geodesic error modulo the half-turn symmetry of a parallel gripper. These conditions specify how closely the transported configuration must match the reference before policy execution resumes.

For placement, we convert the median release offset in the destination frame to an end-effector target using the current grasp offset. The acceptance region $\Omega_d$ intersects the physical destination range with the $Q_{.95}$ release scatter. For destination class $f$, the median profile $m_f$ and its $Q_{.05}$ lower envelope $\underline m_f$ describe the held object's bottom height against horizontal distance to the destination.

\textbf{Approaching the reference.}
ReGuide transports the end-effector with translation and rotation bounded by demonstrated per-step motion. For position error $\Delta_t=p_e^\star-p_t$ and rotation error of unit axis $\omega_t$ and angle $\vartheta_t$, it commands
\begin{equation}
u_t=\frac{\Delta_t}{\|\Delta_t\|}\cdot\frac{\min\{\|\Delta_t\|,\,s\}}{\kappa_p\sigma_p},\qquad
w_t=\omega_t\cdot\frac{\min\{\vartheta_t,\,s_R\}}{\kappa_R\sigma_R},
\label{eq:flow}
\end{equation}
where $s_R$ is the $Q_{.95}$ rotation step and the denominators account for command scale and tracking gain. While carrying, the height target follows $m_f$ and is floored by the intervening crest of $\underline m_f$, following the demonstrated order of lifting before translating. On entering $\mathcal{K}_e$, ReGuide discards the pending chunk and queries the frozen policy on the true observation.

\textbf{Coordinating execution and recovery.}
The same references govern execution after hand-back. Policy actions are constrained by the lateral tolerance $r_e$ during descent and by the height floor during carrying. Contact transitions also determine whether a transport remains applicable. It dissolves on a changed premise, such as a grasp or drop, or when its error stops decreasing over $n$ replans. Each stage permits at most two transports. Completion certificates monitor loss of the destination, release outside $\Omega_d$ and insufficient progress during carrying or placement. They re-arm transport, or complete descent and gripper release at the demonstrated height when the held object remains inside $\Omega_d$ without descending for longer than the $Q_{.95}$ demonstrated dwell. Transport onto a supporting surface uses the same release procedure. Algorithm~\ref{alg:reguide} and Appendix~\ref{app:method} specify the execution protocol.

\begin{figure*}[t]
    \centering
    \includegraphics[width=0.8\textwidth]{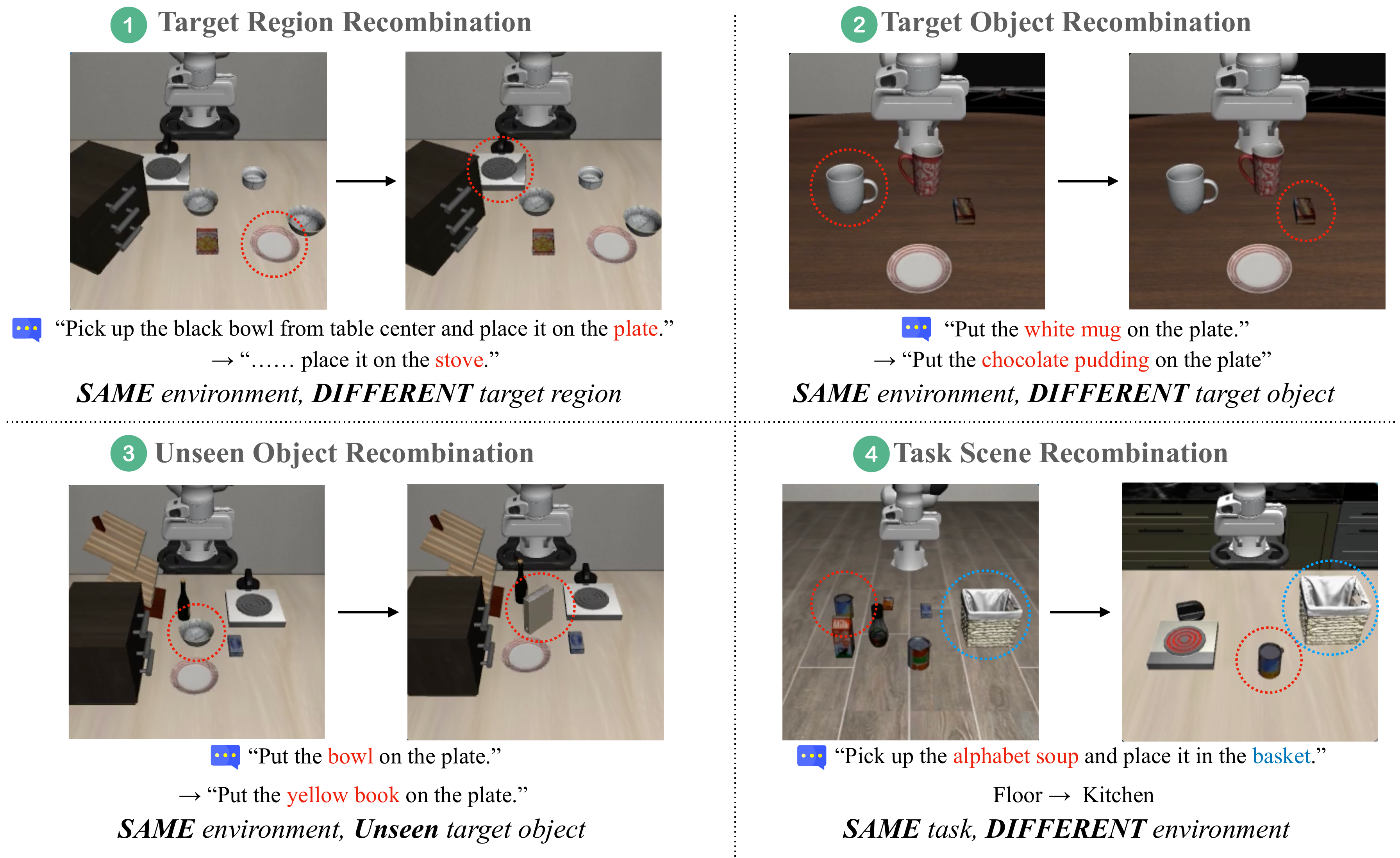}
    \caption{\textbf{Compositional Tasks on LIBERO}. Each setting changes exactly one factor in a training task while keeping the others fixed. Together, they isolate where compositional generalization fails and test whether the policy follows instructions beyond memorized task configurations.}
    \label{fig:libero_shift}
\end{figure*}

\begin{figure*}[t]
    \centering
    \includegraphics[width=0.8\textwidth]{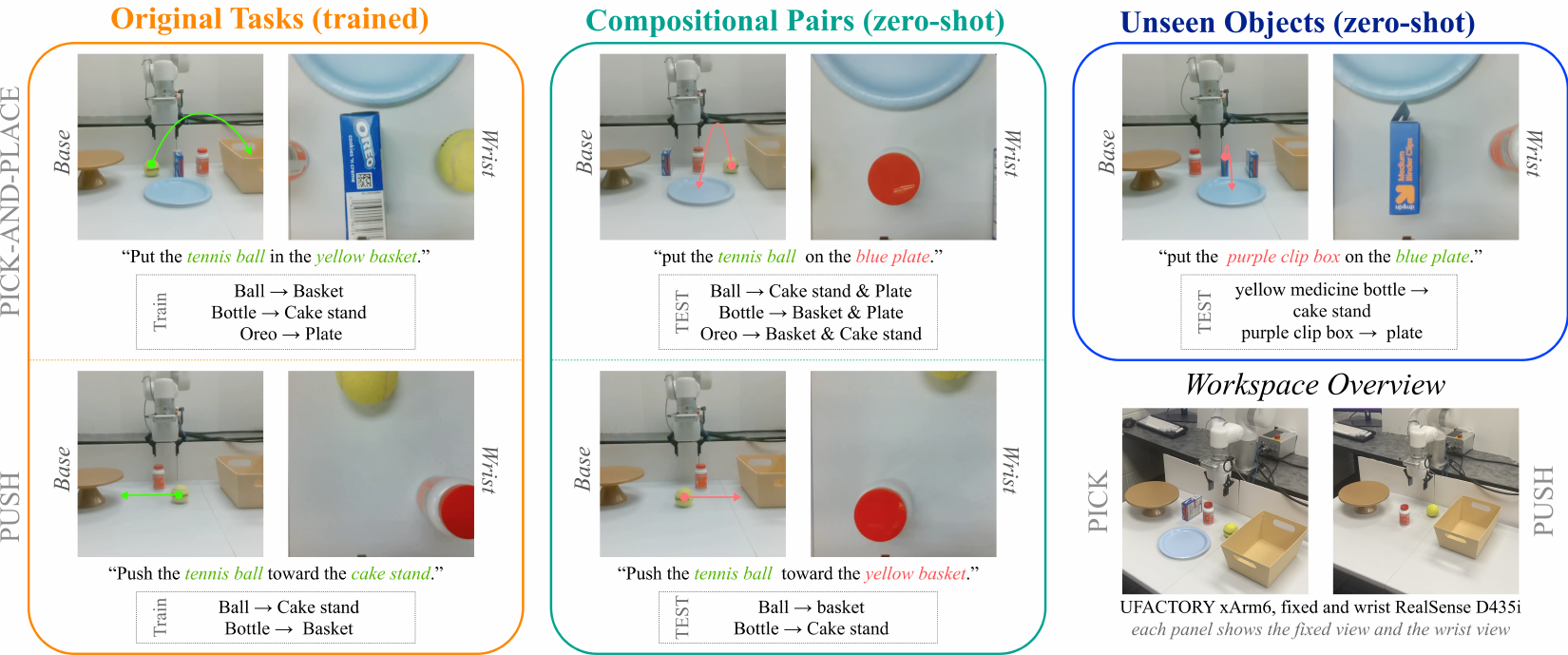}
    \caption{\textbf{Real-robot evaluation on the xArm6}. The setup includes five trained pairings, eight recombined pairings, and two unseen object cases across \textit{pick and place} and \textit{pushing}. Bare $\pi_{0.5}$ and ReGuide are evaluated on the same paired layouts per cell.}
    \label{fig:real_overview}
\end{figure*}

\section{Experiments}

\subsection{Experimental Setup}

\textbf{Simulation Setup.} We build a benchmark on LIBERO that isolates compositional shifts by varying one factor at a time~\citep{liu2023libero} (Fig.~\ref{fig:libero_shift}). We evaluate 4 settings of 4 tasks, yielding 16 composition cells. Each cell modifies a training task along 1 axis: the destination (\textit{Target Region}), the target object (\textit{Target Object}), an object never manipulated in the 4 training suites (\textit{Unseen Object}), or the host scene (\textit{Task Scene}). Another 10 unmodified training tasks form \textit{Orig.} to assess whether the policy retains its original capabilities. All methods replay the same 50 initial states per cell, giving 1300 paired trials each. We evaluate 5 VLA backbones~\citep{kim2025fine,black2026pi0,pi05,zheng2026xvla,nvidia2025gr00t,nvidia2026gr00tn17} and 4 methods that strengthen grounding or instruction following~\citep{huang2025otter,GuidedVLA,xu2026apt,fang2026vision}. We wrap 3 backbones with ReGuide using the same demonstration statistics and oracle simulator state to assess the performance ceiling (Appendix~\ref{app:bench}). We also evaluate perception-based grounding.

\begin{table*}[t!]
\centering
\caption{\textbf{Success rate (\%) on the single-factor composition evaluation of LIBERO.} Each axis recombines one factor of a training task and pools four cells $\times$ 50 paired trials. \textit{Orig.} reports the ten unmodified anchor tasks under the same protocol; \textit{All} pools composition and \textit{Orig.} trials, while \textit{Comp.\ Avg.} pools the 16 composition cells only. Subscripts are 95\% Wilson confidence half-widths.}
\label{tab:composition_main}
\resizebox{\linewidth}{!}{%
\begin{tabular}{c l c c cccc cc}
\toprule
& \multirow{2}{*}{\textbf{Method}} & \multirow{2}{*}{\textbf{Year}} & \multirow{2}{*}{\textbf{Orig.}} & \multicolumn{4}{c}{\textbf{Recombination Axes}} & \multicolumn{2}{c}{\textbf{Average}} \\
\cmidrule(lr){5-8} \cmidrule(lr){9-10}
& & & & \textbf{Target Region} & \textbf{Target Object} & \textbf{Unseen Object} & \textbf{Task Scene} & \textbf{All} & \textbf{Comp.\ Avg.} \\
\midrule
\multirow{5}{*}{\rotatebox[origin=c]{90}{\textit{Found.}}}
& OpenVLA-OFT & 2025 & $\mathbf{97.6}_{\pm1.4}$ & $13.5_{\pm4.7}$ & $1.5_{\pm1.9}$ & $34.5_{\pm6.5}$ & $6.5_{\pm3.5}$ & $46.2_{\pm2.7}$ & $14.0_{\pm2.4}$ \\
& $\pi_0$ & 2025 & $86.0_{\pm3.0}$ & $5.0_{\pm3.1}$ & $1.5_{\pm1.9}$ & $27.0_{\pm6.1}$ & $0.5_{\pm1.3}$ & $38.3_{\pm2.6}$ & $8.5_{\pm1.9}$ \\
& $\pi_{0.5}$ & 2025 & $94.0_{\pm2.1}$ & $40.5_{\pm6.7}$ & $25.0_{\pm6.0}$ & $49.5_{\pm6.9}$ & $41.0_{\pm6.8}$ & $60.2_{\pm2.7}$ & $39.0_{\pm3.4}$ \\
& X-VLA & 2026 & $\mathbf{97.6}_{\pm1.4}$ & $39.0_{\pm6.7}$ & $9.5_{\pm4.1}$ & $34.0_{\pm6.5}$ & $0.0_{\pm0.9}$ & $50.2_{\pm2.7}$ & $20.6_{\pm2.8}$ \\
& GR00T N1.7 & 2026 & $94.6_{\pm2.0}$ & $14.5_{\pm4.9}$ & $9.5_{\pm4.1}$ & $28.0_{\pm6.2}$ & $0.0_{\pm0.9}$ & $44.4_{\pm2.7}$ & $13.0_{\pm2.3}$ \\
\midrule
\multirow{4}{*}{\rotatebox[origin=c]{90}{\textit{Comp.}}}
& OTTER & 2025 & $82.2_{\pm3.3}$ & $0.0_{\pm0.9}$ & $0.0_{\pm0.9}$ & $13.0_{\pm4.7}$ & $0.5_{\pm1.3}$ & $33.7_{\pm2.6}$ & $3.4_{\pm1.3}$ \\
& GuidedVLA & 2026 & $94.8_{\pm2.0}$ & $2.0_{\pm2.1}$ & $6.5_{\pm3.5}$ & $26.5_{\pm6.1}$ & $0.5_{\pm1.3}$ & $41.9_{\pm2.7}$ & $8.9_{\pm2.0}$ \\
& APT & 2026 & $91.4_{\pm2.5}$ & $37.0_{\pm6.6}$ & $33.5_{\pm6.5}$ & $25.5_{\pm6.0}$ & $1.5_{\pm1.9}$ & $50.2_{\pm2.7}$ & $24.4_{\pm3.0}$ \\
& CAG ($\pi_{0.5}$) & 2026 & $81.4_{\pm3.4}$ & $53.0_{\pm6.9}$ & $35.0_{\pm6.6}$ & $44.0_{\pm6.8}$ & $33.5_{\pm6.5}$ & $56.8_{\pm2.7}$ & $41.4_{\pm3.4}$ \\
\midrule
\multirow{3}{*}{\rotatebox[origin=c]{90}{\textit{Ours}}}
& ReGuide (OpenVLA-OFT) & -- & $\underline{96.8}_{\pm1.6}$ & $\underline{87.0}_{\pm4.7}$ & $49.0_{\pm6.9}$ & $\underline{72.5}_{\pm6.1}$ & $74.5_{\pm6.0}$ & $\underline{80.8}_{\pm2.1}$ & $\underline{70.8}_{\pm3.1}$ \\
& ReGuide ($\pi_{0.5}$) & -- & $94.2_{\pm2.1}$ & $\mathbf{96.0}_{\pm2.8}$ & $\mathbf{93.0}_{\pm3.6}$ & $\mathbf{94.5}_{\pm3.2}$ & $\underline{94.5}_{\pm3.2}$ & $\mathbf{94.4}_{\pm1.3}$ & $\mathbf{94.5}_{\pm1.6}$ \\
& ReGuide (GR00T N1.7) & -- & $94.2_{\pm2.1}$ & $69.0_{\pm6.4}$ & $\underline{52.5}_{\pm6.9}$ & $59.0_{\pm6.8}$ & $\mathbf{96.0}_{\pm2.8}$ & $78.8_{\pm2.2}$ & $69.1_{\pm3.2}$ \\
\bottomrule
\end{tabular}%
}
\end{table*}

\begin{table}[t!]
\centering
\caption{\textbf{Real-robot evaluation on the xArm6.} Entries are successes/trials for trained (\textit{Orig.}), recombined (\textit{Comp.}), and unseen-object (\textit{Unseen}) tasks. Average success rates (\%) pool 50 original, 100 recombined and unseen-object, or all 150 trials per method. Subscripts give 95\% Wilson confidence half-widths. Both methods use the same paired layouts.}
\label{tab:real_main}
\small
\resizebox{\linewidth}{!}{%
\begin{tabular}{l cccccccc}
\toprule
\multirow{2}{*}{\textbf{Method}} & \multicolumn{3}{c}{\textbf{Pick-and-Place}} & \multicolumn{2}{c}{\textbf{Push}} & \multicolumn{3}{c}{\textbf{Average (\%)}} \\
\cmidrule(lr){2-4} \cmidrule(lr){5-6} \cmidrule(lr){7-9}
& \makebox[1.3cm][c]{\textbf{Orig.}} & \makebox[1.3cm][c]{\textbf{Comp.}} & \makebox[1.3cm][c]{\textbf{Unseen}} & \makebox[1.3cm][c]{\textbf{Orig.}} & \makebox[1.3cm][c]{\textbf{Comp.}} & \makebox[1.3cm][c]{\textbf{Orig.}} & \makebox[1.3cm][c]{\textbf{Comp.}} & \makebox[1.3cm][c]{\textbf{All}} \\
\midrule
Bare $\pi_{0.5}$   & \makebox[1.3cm][c]{26/30} & \makebox[1.3cm][c]{7/60} & \makebox[1.3cm][c]{8/20} & \makebox[1.3cm][c]{17/20} & \makebox[1.3cm][c]{0/20} & \makebox[1.3cm][c]{$86.0_{\pm9.6}$} & \makebox[1.3cm][c]{$15.0_{\pm7.0}$} & \makebox[1.3cm][c]{$38.7_{\pm7.7}$} \\
\textbf{+ ReGuide} & \makebox[1.3cm][c]{28/30} & \makebox[1.3cm][c]{53/60} & \makebox[1.3cm][c]{17/20} & \makebox[1.3cm][c]{18/20} & \makebox[1.3cm][c]{20/20} & \makebox[1.3cm][c]{$92.0_{\pm7.8}$} & \makebox[1.3cm][c]{$90.0_{\pm6.0}$} & \makebox[1.3cm][c]{$90.7_{\pm4.7}$} \\
\bottomrule
\end{tabular}%
}
\end{table}

\textbf{Real-World Setup.} We fine-tune $\pi_{0.5}$ on 250 scripted demonstrations of 3 pick-and-place and 2 push tasks on a UFACTORY xArm6 with base and wrist RGB-D cameras. The target objects and destinations vary in shape and size. The 15 evaluation cells cover the 5 training pairings, 8 recombined pairings and two with an unseen object, each under 10 layouts shared by the bare policy and ReGuide, that is 150 paired trials per method. Grounding uses an open-vocabulary detector with depth and measured object dimensions to assess real-world performance (Appendix~\ref{app:real}).

\subsection{Experimental Results and In-Depth Analysis}

\begin{figure}[t]
\centering
\includegraphics[width=0.7\linewidth]{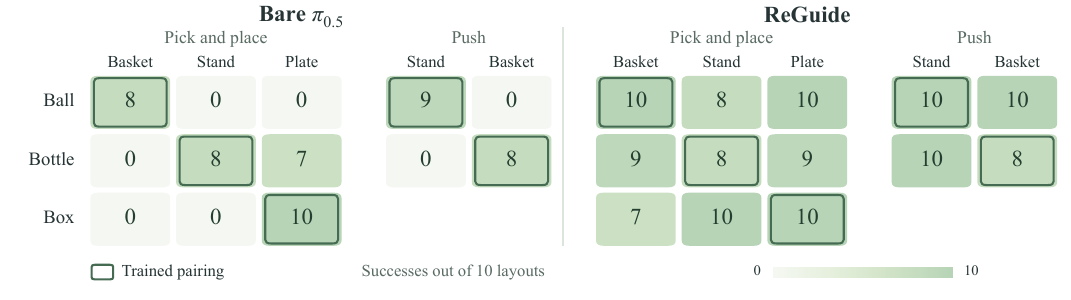}
\caption{\textbf{Real-robot success per object and destination pairing.} Off the trained diagonal the bare policy succeeds on one pairing (7/10); ReGuide reaches at least 7/10 on every pairing.}
\label{fig:real_grid}
\end{figure}

\begin{figure}[t!]
\centering
\includegraphics[width=0.8\linewidth]{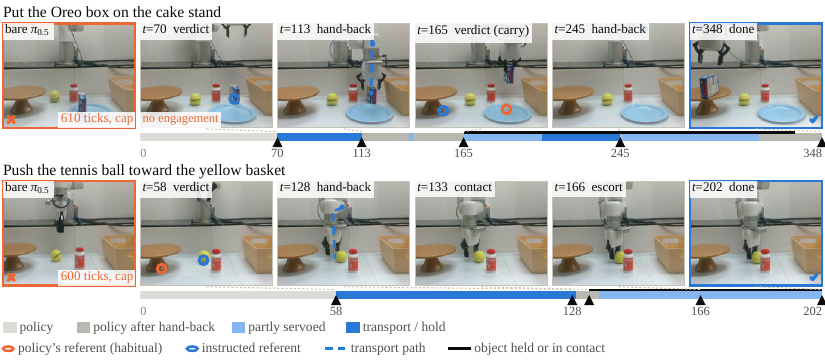}
\caption{\textbf{Paired rollouts under recomposition.} The bare policy acts on the habitual referent, while ReGuide transfers the interaction configuration to the instructed referent and guides execution there.}
\label{fig:qual}
\end{figure}

\textbf{ReGuide recovers most of the loss caused by recomposition.}
The comparison methods in Table~\ref{tab:composition_main} strengthen instruction conditioning but do not explicitly rebind execution to demonstration-derived interaction configurations. Their successes concentrate in the LIBERO-Goal scene, trained under 10 instructions (187/195 for APT), or in cells where training behavior already satisfies the new instruction (50/71 for GuidedVLA). CAG improves \textit{Target Region} and \textit{Target Object} by 12.5 and 10.0 points, but raises \textit{Comp.\ Avg.} only from 39.0 to 41.4 while reducing \textit{Orig.} from 94.0 to 81.4. Without ReGuide, \textit{Task Scene} remains at or below 6.5\% for all non-$\pi_{0.5}$ policies despite unchanged training instructions. ReGuide improves \textit{Comp.\ Avg.} by 55.5--56.8 points across backbones, with \textit{Orig.} changing by at most 0.8. Remaining failures on OpenVLA-OFT and GR00T N1.7 concentrate after hand-back, notably on the moka pot (2/50 each versus 49/50 for $\pi_{0.5}$), consistent with differences in local competence after reaching the interaction configuration (\eqref{eq:factor}). With perception-based grounding, our strongest backbone, $\pi_{0.5}$, achieves 87.5\% compositional success, 7.0 points below oracle grounding while retaining a 48.5-point gain over the bare policy (Appendix~\ref{app:perception_grounding}). On the xArm6, ReGuide raises \textit{Comp.} success from 15.0\% to 90.0\%, with 75 paired wins and no losses, while \textit{Orig.} improves from 86.0\% to 92.0\% (Table~\ref{tab:real_main}). Gains span recombined pairings and unseen objects (Figs.~\ref{fig:real_grid} and~\ref{fig:qual}).

\textbf{Local interaction skills remain usable under recomposition.}
We transport the end-effector to the instructed object's demonstration-derived pre-contact target, then run frozen $\pi_{0.5}$ without further guidance. Over 50 paired trials per cell, grasp success rises from 64.5\% to 95.8\% on 8 recomposed cells, supporting access to local grasping through configuration transfer (Appendix~\ref{app:precontact_diag}).

\textbf{The policy's own actions matter after hand-back.}
 To test execution from these configurations, we replace VLA actions after the first hand-back while retaining the wrapper. Over 800 paired composition trials, success falls from 94.5\% to 38.1\% with a default descent-and-close action and 28.0\% with a hold action (exact McNemar, both $p<10^{-100}$). Grasp success falls from 98.6\% to 44.0\% and 33.5\%, respectively (Appendix~\ref{app:handback_replacement}).

\begin{table}[t!]
\centering
\caption{\textbf{Component-group ablations and hand-back assistance (success, \%).} The first 4 removals test semantic and geometric rebinding; the final removal isolates post-hand-back assistance.}
\label{tab:ablation_mech}
\resizebox{\linewidth}{!}{%
\begin{tabular}{l cccc c c}
\toprule
\multirow{2}{*}{\textbf{Variant}} & \multicolumn{4}{c}{\textbf{Recombination Axes}} & \multirow{2}{*}{\textbf{Comp.\ Avg.}} & \multirow{2}{*}{$\boldsymbol{\Delta}$ \textbf{vs Full}} \\
\cmidrule(lr){2-5}
& \textbf{Target Region} & \textbf{Target Object} & \textbf{Unseen Object} & \textbf{Task Scene} & & \\
\midrule
Bare $\pi_{0.5}$ & $40.5_{\pm6.7}$ & $25.0_{\pm6.0}$ & $49.5_{\pm6.9}$ & $41.0_{\pm6.8}$ & $39.0_{\pm3.4}$ & $-55.5^{***}$ \\
Naive transport & $43.5_{\pm6.8}$ & $70.5_{\pm6.3}$ & $75.5_{\pm5.9}$ & $84.5_{\pm5.0}$ & $68.5_{\pm3.2}$ & $-26.0^{***}$ \\
\midrule
w/o competing referents & $86.5_{\pm4.7}$ & $77.0_{\pm5.8}$ & $\underline{91.5}_{\pm3.9}$ & $88.5_{\pm4.4}$ & $85.9_{\pm2.4}$ & $-8.6^{***}$ \\
w/o chunk commitment & $89.0_{\pm4.4}$ & $83.5_{\pm5.1}$ & $81.5_{\pm5.4}$ & $\underline{91.5}_{\pm3.9}$ & $86.4_{\pm2.4}$ & $-8.1^{***}$ \\
w/o pre-contact sets & $87.0_{\pm4.7}$ & $70.5_{\pm6.3}$ & $75.5_{\pm5.9}$ & $70.5_{\pm6.3}$ & $75.9_{\pm3.0}$ & $-18.6^{***}$ \\
w/o bounded transport & $84.0_{\pm5.1}$ & $\underline{87.5}_{\pm4.6}$ & $90.0_{\pm4.2}$ & $89.0_{\pm4.4}$ & $\underline{87.6}_{\pm2.3}$ & $-6.9^{***}$ \\
\midrule
w/o hand-back assistance & $\underline{94.0}_{\pm3.4}$ & $76.5_{\pm5.8}$ & $86.5_{\pm4.7}$ & $81.5_{\pm5.4}$ & $84.6_{\pm2.5}$ & $-9.9^{***}$ \\
\midrule
\textbf{ReGuide (full)} & $\mathbf{96.0}_{\pm2.8}$ & $\mathbf{93.0}_{\pm3.6}$ & $\mathbf{94.5}_{\pm3.2}$ & $\mathbf{94.5}_{\pm3.2}$ & $\mathbf{94.5}_{\pm1.6}$ & -- \\
\bottomrule
\multicolumn{7}{l}{\footnotesize $\Delta$: paired difference to Full in pp over the same 800 trials; McNemar exact test, $^{*}p<0.05$, $^{**}p<0.01$, $^{***}p<0.001$.} \\
\end{tabular}%
}
\end{table}

\begin{figure}[t]
\centering
\includegraphics[width=0.8\linewidth]{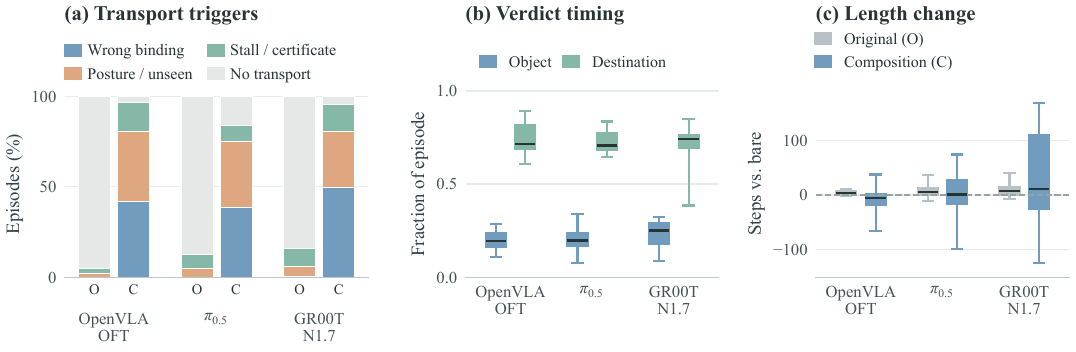}
\caption{\textbf{When, how often and at what cost ReGuide intervenes.} Classes are the verdicts of ReGuide, excluding one-step transports cancelled at a clause switch.}
\label{fig:when}
\end{figure}

\textbf{Guidance tracks referents and interaction progress.}
In simulation, transport occurs in 83.9--96.5\% of composition versus 5.0--16.2\% of original episodes (Fig.~\ref{fig:when}). Only 3/1500 original episodes trigger wrong-binding corrections. Other interventions establish compatible posture, recover stalled approaches, or support completion. Among jointly successful trials, median length changes are $-5$ to $+11$ steps on composition tasks and $+4$ to $+7$ on original tasks. Transport occupies a median 14--15\% of control steps in successful composition episodes with interventions (Appendix~\ref{app:intervention_stats}).

\textbf{Comparison with agent-based guidance and runtime.}
On one paired initial state per cell, ReGuide solves 23/26 tasks and Harness VLA~\citep{zhang2026harness} solves 18/26, an exploratory comparison reported separately from Table~\ref{tab:composition_main}. ReGuide's median replanning latency is 129.2\,ms versus 129.0\,ms for the bare policy, while Harness incurs a median 6.1\,s wait per planner response. Beyond latency, Harness incurs an additional GPT-5.5 API cost averaging \$1.84 per episode (Appendix~\ref{app:harness_runtime}).

\begin{figure}[!t]
\centering
\includegraphics[width=0.8\linewidth]{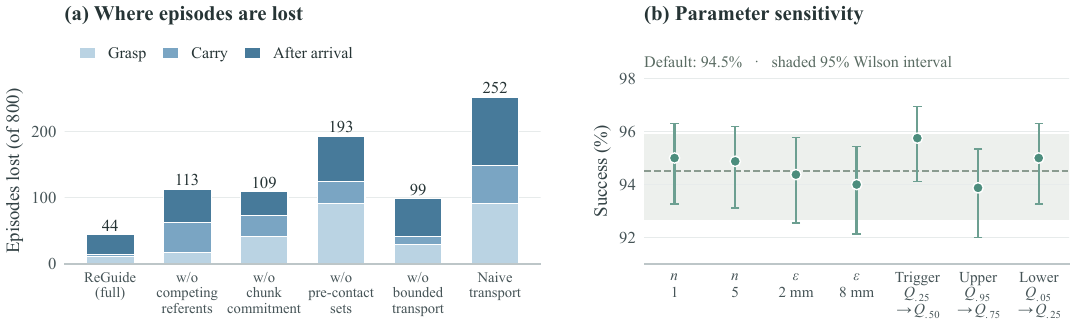}
\caption{\textbf{Failure stages and parameter sensitivity.} (a) Failure stage by removed component. (b) No tested parameter variant differs significantly from the default under paired McNemar tests.}
\label{fig:ablation_fig}
\end{figure}

\subsection{Ablation Studies}

\textbf{Component groups and hand-back assistance.}
Table~\ref{tab:ablation_mech} tests referent monitoring, interaction configurations, and execution coordination. Naive transport recovers 29.5 of the 55.5-point gain by servoing to a fixed height above the grounded object and destination. Removing pre-contact sets or the chunk commitment group mainly impairs grasping, removing competing referents impairs destination selection, and removing the bounded transport group increases post-arrival failures (Fig.~\ref{fig:ablation_fig}a). Without pre-contact sets, objects held fall from 789/800 to 708/800. Without competing referents, destination arrivals fall from 785/800 to 737/800, as habitual destinations are excluded. Disabling post-hand-back constraints and completion assistance while retaining subsequent transports yields 84.6\% success and 97.1\% grasp success, showing the contribution of execution support after configuration transfer (Appendix~\ref{app:ablation}). Stability across tested constants and quantile levels supports the framework's shared settings without precise tuning (Fig.~\ref{fig:ablation_fig}b).

\section{Conclusion}

Under recomposition, a substantial fraction of VLA failures arise from incorrect referent binding, while local manipulation competence often remains recoverable from demonstration-derived configurations. ReGuide combines semantic and geometric rebinding to restore referent alignment and coordinate guidance, local policy execution, and recovery using demonstration-derived configurations and motion bounds. Without retraining the policy, ReGuide raises compositional success by 55.5 to 56.8 percentage points across three backbones in simulation and from 15.0\% to 90.0\% on a real robot, while maintaining comparable performance on original tasks.

\bibliography{iclr2027_conference}
\bibliographystyle{iclr2027_conference}

\appendix
\setcounter{topnumber}{4}\setcounter{bottomnumber}{3}\setcounter{totalnumber}{8}
\renewcommand{\topfraction}{0.92}\renewcommand{\bottomfraction}{0.85}\renewcommand{\textfraction}{0.05}\renewcommand{\floatpagefraction}{0.75}

\providecommand{\TODO}[1]{\textcolor{red}{[TODO #1]}}
\newcolumntype{L}[1]{>{\raggedright\arraybackslash}p{#1}}
\makeatletter\@ifundefined{fitbox}{\newsavebox{\fitbox}}{}\makeatother
\providecommand{\fitwidth}[1]{\sbox{\fitbox}{#1}\ifdim\wd\fitbox>\linewidth\resizebox{\linewidth}{!}{\usebox{\fitbox}}\else\usebox{\fitbox}\fi}

\section{Method Details}
\label{app:method}

\subsection{Notation}

An action chunk is $A_t=(a_t^1,\dots,a_t^H)$, with $h\le H$ actions executed before replanning. The end-effector pose is $x_t=(p_t,R_t)\in SE(3)$, with translation and rotation-vector commands in $\mathbb{R}^3$. Tracking gains satisfy $\kappa_p,\kappa_R\in(0,1]$.

\begin{table}[htbp]
\centering
\caption{Notation used in the method and its implementation details.}
\label{tab:notation}
\small
\begin{tabular}{@{}L{1.60in}L{3.70in}@{}}
\toprule
\textbf{Symbol} & \textbf{Meaning} \\
\midrule
$\pi_\theta$, $A_t$, $H$, $h$ & frozen policy, action chunk, chunk length, executed actions per replan \\
$a=(u,w,g)$, $x_t=(p_t,R_t)$ & translation, rotation and gripper command, end-effector pose \\
$\sigma$, $\kappa$ & command scales of the controller, tracking gains fitted on demonstrations \\
$e$, $c_e$, $R_e$, $b_e$ & scene entity, its position, orientation and bounding-box half-extents \\
$e_k^\star$, $q_t$ & instructed referent of stage $k$, moving point (end-effector or held object) \\
$\mathcal{R}_t$, $\hat q_t$, $\hat R_t$ & unexecuted chunk remainder, predicted endpoint and posture \\
$\delta_e$, $\rho_e$, $\chi_t(e)$, $e_t^\pi$ & offset from the box of $e$, commitment radius, commitment indicator, policy's referent \\
$\mathcal{H}_k$, $\mathcal{W}$, $\mathcal{E}_k$ & habitual referents, remaining movable entities, candidate set \\
$C_e$, $t_e$, $(\bar o,\bar R)$, $\bar z$ & canonical alignment, top-face centre, grasp offset and rotation medoid, entry altitude \\
$p_e^\star$, $R_e^\star$, $\mathcal{K}_e$ & pre-contact target and hand-back tolerance set \\
$s$, $s_R$, $\tau_R$, $r_e$ & step bounds, posture tolerance, lateral tolerance of the descent \\
$m_f$, $\underline{m}_f$, $\Omega_d$ & median height profile of class $f$, its lower envelope, acceptance region of $d$ \\
$\phi(e)$ & shape descriptor (height, two widths, top fill) \\
$Q_\alpha$, $n$, $N$, $\varepsilon$ & empirical quantile, consecutive count, failure-to-bind window, position resolution \\
$\beta_k$, $\gamma_k$, $\tilde\gamma_k$, $\nu_k$, $\eta_k$ & binding, competence, competence after a transport, false and missed interventions \\
\bottomrule
\end{tabular}
\end{table}

\subsection{Algorithm}

\begin{algorithm}[htbp]
\caption{ReGuide control loop for one stage $k$}
\label{alg:reguide}
\begin{algorithmic}[1]
\Require frozen policy $\pi_\theta$, instruction $\ell$, grounding module, demonstration statistics
\State budget $\gets 2$, transport $\gets$ none, $\mathcal{R}_t \gets \emptyset$
\For{each control step $t$}
  \State read $x_t$ and the entities of the scene, update the interaction state and the moving point $q_t$
  \If{a transport is active}
    \If{its premise changed \textbf{or} its error did not decrease over $n$ replans}
      \State dissolve the transport
    \ElsIf{$x_t\in\mathcal{K}_e$ (or the held object is one step above the release height over $d$)}
      \State end the transport, discard $\mathcal{R}_t$ \Comment{hand-back on the true observation}
    \Else
      \State execute the bounded command of \eqref{eq:flow}, with the height following $m_f$ while an object is held
      \State \textbf{continue}
    \EndIf
  \EndIf
  \If{$\mathcal{R}_t=\emptyset$} $A_t \gets \pi_\theta(o_t,\ell)$, $\mathcal{R}_t \gets A_t$ \EndIf
  \State compute $\hat q_t$, $\hat R_t$ (Eq.~\ref{eq:endpoint}) and the policy's referent $e_t^\pi$ (Eq.~\ref{eq:commit})
  \If{budget $>0$ \textbf{and} (wrong binding \textbf{or} posture mismatch \textbf{or} failure to bind \textbf{or} unseen object at the first decision)}
    \State set the target of \eqref{eq:target} or the destination target, start a transport, budget $\gets$ budget $-1$
  \Else
    \State pop the next action of $\mathcal{R}_t$, clamp it to the demonstrated envelope, execute it
  \EndIf
  \If{a completion certificate holds} re-arm the transport or complete the release \EndIf
\EndFor
\end{algorithmic}
\end{algorithm}

Algorithm~\ref{alg:reguide} links referent monitoring to configuration transfer. It checks predicted commitment and approach progress, transports toward the selected reference when needed, and resumes policy execution from the true observation. Motion envelopes and completion conditions coordinate the subsequent interaction.

\subsection{Binding and Competence Under Intervention}
\label{app:binding_analysis}

A wrapper intervenes on a correct binding with probability $\nu_k$, misses a wrong binding with probability $\eta_k$, and on every intervention delivers a state with competence $\tilde\gamma_k$. If intervention and outcome are independent given binding, then
\begin{equation}
\Pr\nolimits_{\mathrm{RG}}(S_k)=\beta_k\big[(1-\nu_k)\,\gamma_k+\nu_k\,\tilde\gamma_k\big]+(1-\beta_k)(1-\eta_k)\,\tilde\gamma_k .
\label{eq:budget}
\end{equation}

\textit{Derivation of \eqref{eq:budget}.} Let $I_k$ be the event that the wrapper intervenes in stage $k$. Partitioning on $B_k$ and $I_k$ gives
$\Pr(S_k)=\Pr(B_k)\big[\Pr(\lnot I_k\mid B_k)\Pr(S_k\mid B_k,\lnot I_k)+\Pr(I_k\mid B_k)\Pr(S_k\mid B_k,I_k)\big]+\Pr(\lnot B_k)\Pr(I_k\mid \lnot B_k)\Pr(S_k\mid \lnot B_k,I_k)$,
where the missing term vanishes because $S_k\subseteq B_k$ under the bare policy, so a wrong binding that is not corrected cannot succeed. With $\nu_k=\Pr(I_k\mid B_k)$, $\eta_k=\Pr(\lnot I_k\mid\lnot B_k)$, $\Pr(S_k\mid B_k,\lnot I_k)=\gamma_k$ by the independence assumption, and $\Pr(S_k\mid\cdot,I_k)=\tilde\gamma_k$, \eqref{eq:budget} follows. Setting $\nu_k=\eta_k=0$ and $\tilde\gamma_k=\gamma_k$ gives $\Pr_{\mathrm{RG}}(S_k)=\gamma_k$, the nominal competence level motivating restoration of the interaction conditions.

\subsection{Stages, Events and Interaction States}
\label{app:events}

Each relocation clause specifies an object and a destination, with an approach stage followed by a carry stage (Sec.~\ref{sec:pf}). Contact events determine which referent, configuration and motion statistics apply.

\textbf{Demonstration events.} Engagement is the first frame from which the entity moves with the end-effector over a window of eight frames. The effect starts when the entity rises more than 3\,cm above its rest height, and termination is the end of co-motion. These events use relative motion without reading gripper width. Approach statistics precede engagement, while carry statistics span effect onset to termination.

\textbf{Execution state.} The same detector tracks five states, free, near, attached, effecting and detaching. The held predicate corresponds to effecting. Posture compatibility is checked only while free, and the failure-to-bind counter runs only without physical interaction with the instructed object. In multi-clause tasks, the active instance switches when the policy commits to or touches an object of another unsatisfied clause. Such an object is a legitimate referent. After a completed clause, the next instance is armed once the end-effector leaves the previous object.

\subsection{Details of Semantic Rebinding}
\label{app:semantic}

\textbf{Competing referents.} The first frame after the scene has settled is embedded with an ImageNet-pretrained ResNet-50 and matched to the first frames of the training demonstrations. The goal predicates of the training tasks of the retrieved scene provide the habitual objects and destinations, and every relocation clause of a habitual task contributes. A habitual object is grounded in the current scene by name when an entity of that name exists and otherwise by geometric identity (Appendix~\ref{app:geo}). Objects of the instruction itself are never habitual referents. The watch set $\mathcal{W}$ contains every other body with a free, sliding or hinge joint.

\textbf{Commitment radii.} The entry event of a demonstration is the first frame at which the end-effector lies inside the bounding box of the entity along two of the three world axes, which covers top grasps and side grasps alike. The commitment radius $\rho_e$ is the $Q_{.25}$ of the distance to the box surface at that frame for the training type of $e$. Typical values in LIBERO are 3.3\,cm for the bowl, 7.8\,cm for the moka pot, 7.2\,cm for the mug, 1.1\,cm for the wine bottle and 1.8\,cm for the milk carton.

\textbf{Referent monitoring during carrying.} Equation~\ref{eq:commit} uses the held object's predicted horizontal position relative to each destination. A mismatch requires commitment to a habitual destination, by heading or dwelling, without commitment to the instructed destination. Heading must persist for $n=3$ consecutive steps, while dwelling counts immediately. Transport also requires a stable grasp and compatibility with the carry envelope. The gripper width must remain at rest for three steps within the demonstrated band, and the held object must lie inside the demonstrated height envelope. Otherwise, the height floor first lifts it into that envelope.

\textbf{Posture compatibility.} Before descent beside the object and while the interaction state is free, guidance is triggered when $\vartheta(R_e^\star,\hat R_t)>\tau_R$ and $\vartheta(R_e^\star,\hat R_t)\ge\vartheta(R_e^\star,R_t)-s_R$. The predicted error must therefore exceed the demonstrated tolerance without improving over the current error by more than one rotation step.

\textbf{Progress toward the grasp configuration.} Progress is measured on three channels, the position residual to the demonstrated grasp point in units of $s$, the posture residual in units of $s_R$, and the closing of the gripper. The window is $N=40$ steps, one more than the longest run without progress in the approach segments of the single-clause demonstrations, and $N=75$ for an instance issued after a completed clause.

\subsection{Details of Geometric Rebinding}
\label{app:geo}

\begin{figure}[htbp]
\centering
\includegraphics[width=\linewidth]{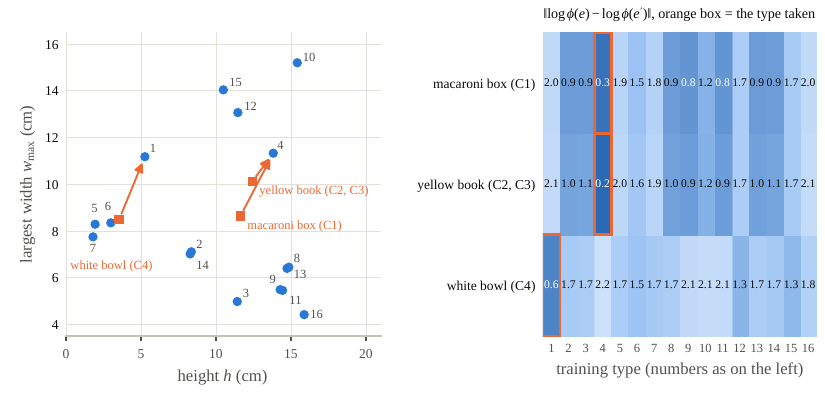}
\caption{Geometric identity. Left, height and largest width of the 16 training types (1 black bowl, 2 alphabet soup, 3 bbq sauce, 4 black book, 5 butter, 6 chocolate pudding, 7 cream cheese, 8 ketchup, 9 milk, 10 moka pot, 11 orange juice, 12 porcelain mug, 13 salad dressing, 14 tomato sauce, 15 white and yellow mug, 16 wine bottle) and of the three unseen objects of the benchmark, with an arrow to the type whose statistics they take. Right, the distance $\|\log\phi(e)-\log\phi(e^\prime)\|$ over the four components of the descriptor between each unseen object and every type. The macaroni and cheese box and the yellow book take the black book, at 0.33 and 0.19, and the white bowl takes the black bowl, at 0.58.}
\label{fig:shapes}
\end{figure}

\begin{figure}[htbp]
\centering
\includegraphics[width=\linewidth]{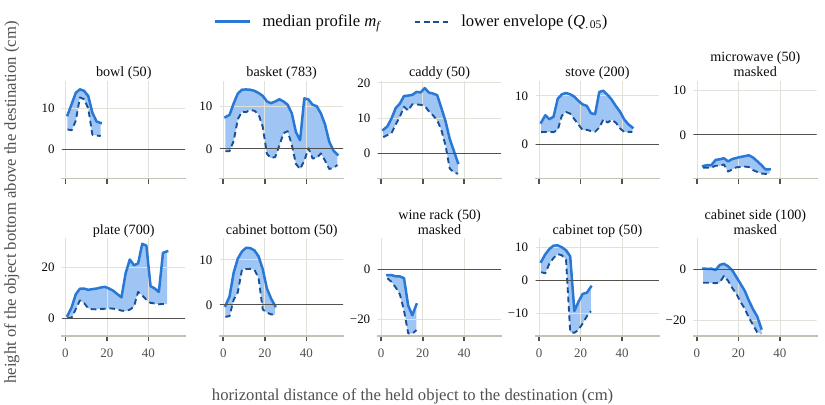}
\caption{Height profiles of the ten destination classes. The height of the bottom of the held object above the destination plane is plotted against the horizontal distance to the destination, as the median $m_f$ (solid) and the lower envelope $\underline{m}_f$, the $Q_{.05}$ (dashed), in bins of 2\,cm with at least 20 samples. Numbers in parentheses are the contributing demonstrations. In the three masked classes the envelope stays below the destination plane over the final approach, and a transport onto them keeps the current height.}
\label{fig:height_profiles}
\end{figure}

\textbf{Canonical frame and geometric identity.} The alignment $C_e$ defines the gravity-aligned frame $R_eC_e$ of \eqref{eq:target}, ordering horizontal axes by extent at rest. The descriptor $\phi(e)$ holds the height, the two widths and the top fill, which is the fraction of a $24\times24$ grid of downward rays whose first hit lies in the top quarter of the object and separates hollow from solid objects. An object is of a training type when its three sorted extents agree with those of the type within $\varepsilon$, and it is unseen otherwise, in which case it takes the statistics of the type with the smallest $\|\log\phi(e)-\log\phi(e')\|$. Names do not enter. A renamed copy of a training object is seen, and a new geometry under a familiar name is unseen. When the footprint of an object is rotationally symmetric within $\varepsilon$, its rotation about the vertical axis is not observable, and the reference configuration uses only the up axis and the top face. Fig.~\ref{fig:shapes} places the 16 training types and the three unseen objects of the benchmark in the descriptor space.

\textbf{Approach configuration.} The target of \eqref{eq:target} hovers above the demonstrated grasp point at the $Q_{.5}$ entry altitude of the type. While the yaw residual exceeds its $Q_{.95}$ tolerance, the hover height is raised to the demonstrated altitude at which the alignment is completed, since the demonstrations align before the final descent. The grasp offset, the rotation medoid and the tolerances are taken from the slice of demonstrations with the same object type and, in order of preference, the same destination, the same destination class, or the class with the closest release tilt.

\textbf{Placement references and carry envelope.} Statistics are pooled by destination class, the declared type of an entity or its interaction region. The median release offset is expressed in the fixture frame and rotated by its yaw. Adding the grasp offset measured at transport onset converts this object reference to an end-effector target. The height profile uses 2\,cm bins with at least 20 samples per bin and 10 demonstrations per class. Its floor is the maximum of $\underline{m}_f$ between the destination and the nearer of the object and the first peak of $m_f$, coordinating lifting with translation. Three of the ten destination classes are approached from the side in LIBERO demonstrations, the microwave, wine rack and cabinet-top side region. Their lower envelope stays below the destination plane, so height-profile transport is masked for these classes. Transport keeps the current height and the policy performs placement. Fig.~\ref{fig:height_profiles} shows the profiles and envelopes.

\textbf{Transport applicability.} A reference is tied to the interaction state at transport onset. The transport dissolves if the held predicate changes, if the object's up axis leaves the rest-frame class used by the demonstrations, or if error fails to decrease by more than $\varepsilon$ over $n$ replans. Each stage permits at most two transports, with the second requiring a completion certificate.

\textbf{Recovery and release.} Completion certificates distinguish loss of the destination from lack of progress toward completion. A transport is re-armed if the held object leaves $\Omega_d$ for $n$ steps after transport, or if it rests outside $\Omega_d$ after release and the policy has disengaged. Recovery also applies when the held object makes no progress toward the destination for $N$ steps, or dwells outside $\Omega_d$ within the carry radius longer than the $Q_{.95}$ demonstrated dwell. If it instead remains inside $\Omega_d$ without descending for longer than the $Q_{.95}$ demonstrated dwell, the wrapper completes descent to the $Q_{.5}$ release height while holding horizontal position at $p_d^\star$, then opens the gripper within the demonstrated release band. Every transport onto a supporting surface uses the same release procedure.

\subsection{Demonstration Statistics and Declared Constants}
\label{app:demo_stats}

\begin{figure}[htbp]
\centering
\includegraphics[width=\linewidth]{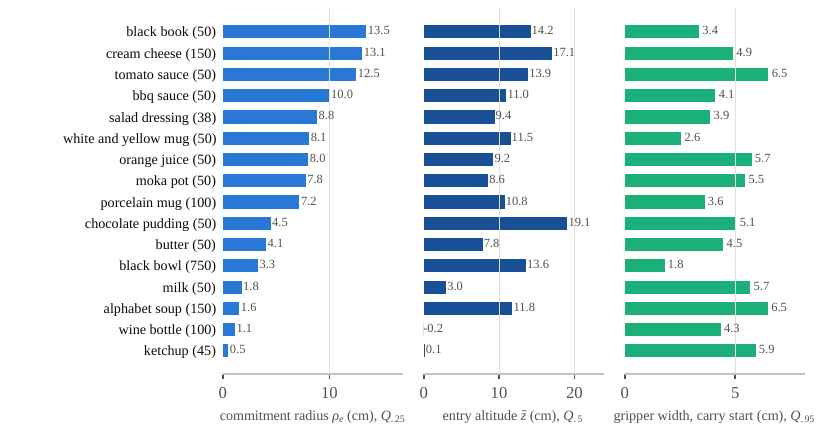}
\caption{Per-type demonstration statistics of the LIBERO training suites. The commitment radius $\rho_e$ is the $Q_{.25}$ of the distance to the box surface at the entry event, the entry altitude $\bar z$ the $Q_{.5}$ of the height of the end-effector above the object top at the entry event, and the gripper band the $Q_{.95}$ of the gripper width at the start of a carry. Numbers in parentheses are the demonstrations of the type. The entry altitude is close to zero for the wine bottle and the ketchup.}
\label{fig:demo_types}
\end{figure}

\begin{table}[htbp]
\centering
\caption{Demonstration statistics read by ReGuide. A slice is a pair of object type and task.}
\label{tab:stats}
\footnotesize
\begin{tabular}{@{}L{1.45in}L{2.30in}L{1.40in}@{}}
\toprule
\textbf{Statistic} & \textbf{Measured on} & \textbf{Level and role} \\
\midrule
tracking gains $\kappa_p$, $\kappa_R$ & realized against commanded motion, per demonstration & $Q_{.5}$, unit conversion \\
step bounds $s$, $s_R$ & per-step displacement and rotation, approach and carry & $Q_{.95}$, upper bound \\
commitment radius $\rho_e$ & distance to the box at the entry event, per object type & $Q_{.25}$, trigger \\
carry radius & distance to the destination during demonstrated carries, per class & $Q_{.25}$, trigger \\
entry altitude $\bar z$ & height above the object top at the entry event, per type & $Q_{.5}$, target \\
grasp offset $\bar o$, rotation $\bar R$ & end-effector pose at engagement in the canonical frame, per slice & median and medoid, target \\
posture tolerance $\tau_R$, yaw tolerance & rotation residual at the entry event, per slice & $Q_{.95}$, upper bound \\
lateral tolerance $r_e$ & scatter of the grasp along the closing axis, per slice & $Q_{.95}$, upper bound \\
release offset $\bar o_f$, tolerance & object position at release in the fixture frame, per class & $Q_{.5}$ target, $Q_{.95}$ bound \\
height profile $m_f$, envelope $\underline{m}_f$ & bottom of the held object against distance to the destination & $Q_{.5}$ target, $Q_{.05}$ lower bound \\
dwell and release durations & frames near the destination, frames from release to rest & $Q_{.95}$, upper bound \\
gripper band at the start of a carry & gripper width and its change, per object type & $Q_{.95}$, upper bound \\
shape descriptor $\phi$, frame $C_e$ & geometry of the training object types & identity and transfer \\
\bottomrule
\end{tabular}
\end{table}

Table~\ref{tab:stats} lists the reference statistics used for commitment, configuration transfer and execution. These statistics are computed once from the demonstrations that trained the policy. In LIBERO these are the demonstrations of the four training suites, which give 16 object types and 10 destination classes, the tracking gains are $\kappa_p=0.237$ and $\kappa_R=0.211$, and the step bounds are $s=1.24$\,cm in the approach, $1.34$\,cm in the carry and $s_R=0.027$\,rad. The constants that are declared rather than mined are the consecutive count $n=3$, the position resolution $\varepsilon=4$\,mm, the bin width of 2\,cm, the evidence floors of 20 samples per bin and 10 demonstrations per class, the event thresholds of Sec.~\ref{app:events}, and the budget of two transports per stage. Sec.~\ref{app:ablation} varies $n$, $\varepsilon$ and the quantile levels. Fig.~\ref{fig:demo_types} shows three of the per-type statistics for the 16 object types.

\section{Simulation Benchmark and Protocol}
\label{app:bench}

\textbf{Benchmark design rationale.}
Existing LIBERO extensions perturb tasks along appearance, layout, wording, viewpoint or lighting (LIBERO-PRO~\citep{zhou2026libpro}, LIBERO-Plus~\citep{fei25libero-plus}), or extrapolate a new task from two trained ones (libero-ood~\citep{Confined}, the behaviour-composition tasks of LIBERO-10-R~\citep{wu2026say}). Our cells instead change exactly one factor of one training task: the destination, the object, an object never manipulated in training, or the host scene. A failure in a cell is therefore attributable to that factor, object and destination bindings are probed separately, and recombination is separated from object novelty.

\subsection{Cells}

\begin{figure}[htbp]
\centering
\includegraphics[width=\linewidth]{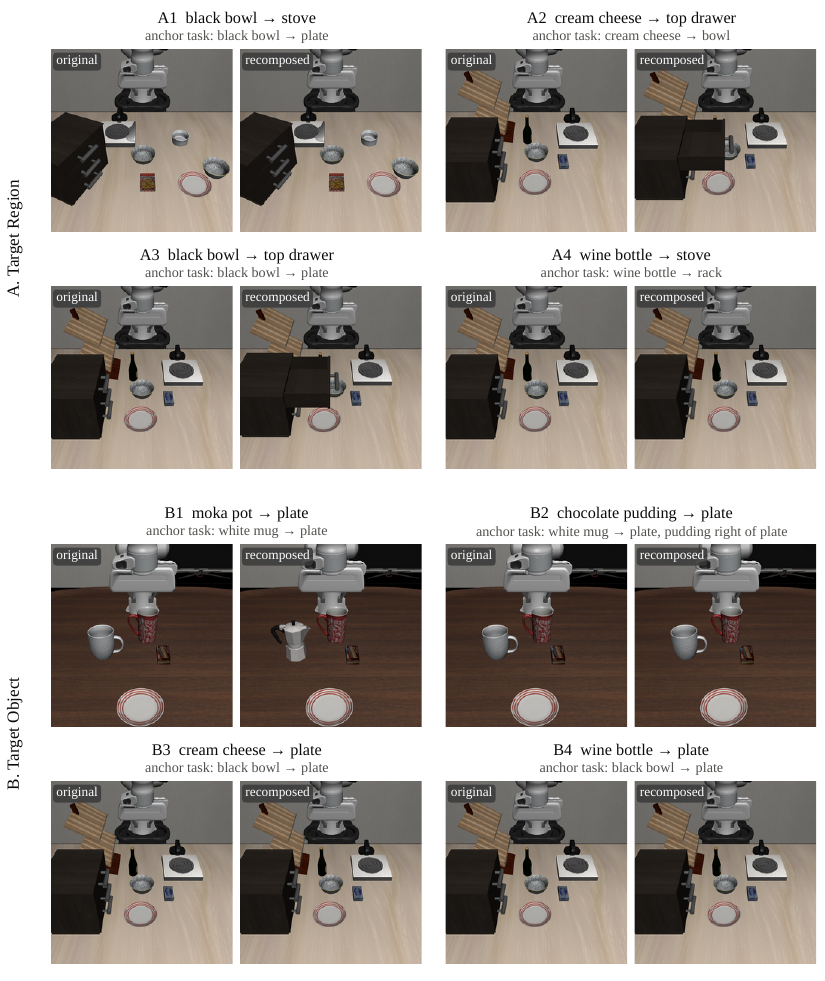}
\caption{Composition cells of the Target Region and Target Object axes. For every cell the left image is the initial state of the unmodified anchor task and the right image the initial state of the cell, rendered from the first initial state of its evaluation bank. The changed factor of each cell is listed in Table~\ref{tab:cells}.}
\label{fig:cells_ab}
\end{figure}

\begin{figure}[htbp]
\centering
\includegraphics[width=\linewidth]{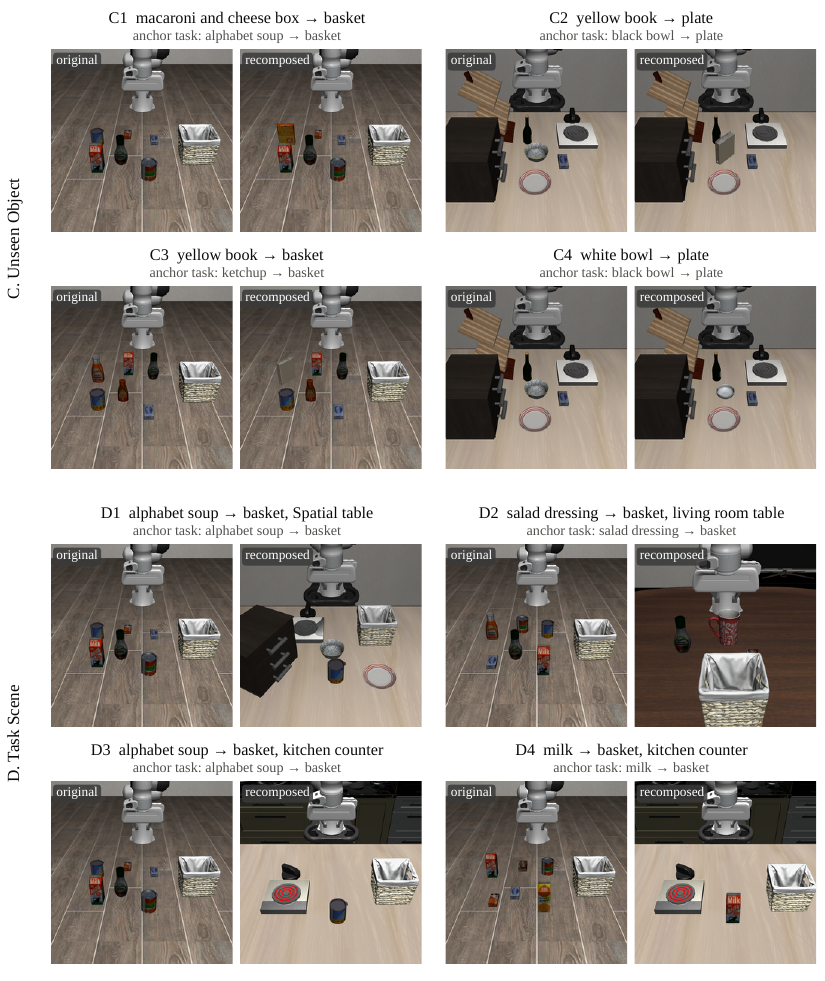}
\caption{Composition cells of the Unseen Object and Task Scene axes, in the layout of Fig.~\ref{fig:cells_ab}. For the Task Scene cells the left image shows the source task in its own scene and the right image the same task on the host surface.}
\label{fig:cells_cd}
\end{figure}

\begin{table}[htbp]
\centering
\caption{The 16 composition cells. Axes A to D are Target Region, Target Object, Unseen Object and Task Scene.}
\label{tab:cells}
\scriptsize
\begin{tabular}{@{}L{0.25in}L{1.45in}L{1.45in}L{1.80in}@{}}
\toprule
\textbf{Cell} & \textbf{Host scene and anchor task} & \textbf{Change} & \textbf{Instruction} \\
\midrule
A1 & Spatial, bowl at table centre to plate & destination is the stove & pick up the black bowl from table center and place it on the stove \\
A2 & Goal, cream cheese to bowl & destination is the open top drawer & put the cream cheese in the top drawer of the wooden cabinet \\
A3 & Goal, bowl to plate & destination is the open top drawer & put the black bowl in the top drawer of the wooden cabinet \\
A4 & Goal, wine bottle to rack & destination is the stove & put the wine bottle on the stove \\
\midrule
B1 & Living room 6, white mug to plate & the mug is replaced in place by a moka pot & put the moka pot on the plate \\
B2 & Living room 6, white mug to plate & the target is the chocolate pudding of the scene & put the chocolate pudding on the plate \\
B3 & Goal, bowl to plate & the target is the cream cheese of the scene & put the cream cheese on the plate \\
B4 & Goal, bowl to plate & the target is the wine bottle of the scene & put the wine bottle on the plate \\
\midrule
C1 & Object, alphabet soup to basket & object is a macaroni and cheese box & pick up the macaroni and cheese box and place it in the basket \\
C2 & Goal, bowl to plate & object is a yellow book & put the yellow book on the plate \\
C3 & Object, ketchup to basket & object is a yellow book & pick up the yellow book and place it in the basket \\
C4 & Goal, bowl to plate & object is a white bowl & put the white bowl on the plate \\
\midrule
D1 & Object, alphabet soup to basket & moved to the Spatial kitchen table & pick up the alphabet soup and place it in the basket \\
D2 & Object, salad dressing to basket & moved to the table of Living room 6 & pick up the salad dressing and place it in the basket \\
D3 & Object, alphabet soup to basket & moved to the counter of Kitchen 8 & pick up the alphabet soup and place it in the basket \\
D4 & Object, milk to basket & moved to the counter of Kitchen 8 & pick up the milk and place it in the basket \\
\bottomrule
\end{tabular}
\end{table}

Every cell is anchored on one training task of LIBERO and changes one factor. The object and the destination of a cell are placed at positions that the arm manipulated in the demonstrations of the host scene, native movable objects that conflict with the new task are removed, and the official region names are reused. We validate scene stability and confirm that the goal predicate is false at initialization and true when the object is placed at the specified goal. Table~\ref{tab:cells} lists the 16 cells, and Figs.~\ref{fig:cells_ab} and \ref{fig:cells_cd} show the initial state of every cell next to its anchor task. The unseen objects are the macaroni and cheese box, which occurs in no LIBERO suite, and the yellow book and the white bowl, which occur in LIBERO-90 but in none of the four suites on which the evaluated checkpoints and the statistics of ReGuide are based. C4 is kept as a reference cell in which the trained behaviour already satisfies the new instruction. The ten original tasks are the anchors of the cells.

\subsection{Protocol}

For every cell, 50 initial simulator states are generated once and stored, and every method replays the same states, so trial $i$ of any two methods starts from the same state. An episode starts with 10 idle steps and ends at success according to the goal predicate of LIBERO or after 520 steps, the largest official budget, for all cells and methods. Table subscripts are 95\% Wilson half-widths. Paired comparisons use the exact two-sided McNemar test on the trials of the two methods with the same initial state.

\subsection{Policies and Comparison Methods}

\begin{table}[htbp]
\centering
\caption{Checkpoints and execution settings.}
\label{tab:policies}
\scriptsize
\begin{tabular}{@{}L{0.85in}L{1.75in}L{2.55in}@{}}
\toprule
\textbf{Method} & \textbf{Checkpoint} & \textbf{Execution} \\
\midrule
$\pi_0$, $\pi_{0.5}$ & official LIBERO checkpoints of openpi & official client, 5 actions executed per replan, chunks of 50 actions for $\pi_0$ and 10 for $\pi_{0.5}$ \\
OpenVLA-OFT & official checkpoint trained on the four suites & chunk of 8 executed open loop, two images and proprioception \\
X-VLA & official LIBERO checkpoint & official client with absolute end-effector targets, the full chunk of 30 actions executed open loop \\
GR00T N1.7 & fine-tuned by us on the four suites, see below & official horizon, 16 actions predicted and 8 executed \\
OTTER & trained by us with the official code, as no checkpoint is released, see below & official evaluation loop, one action per step with a context of 12 frames \\
GuidedVLA & released checkpoint, built on $\pi_0$, see below & official server with corrected depth-adapter checkpoint loading \\
APT & official LIBERO checkpoint & official inference service with an ensemble of 4 \\
CAG & training-free variant on the official $\pi_{0.5}$ checkpoint & authors' server script, guidance weight $\omega=1.5$ as set in their paper for $\pi_{0.5}$ \\
\bottomrule
\end{tabular}
\end{table}

\textbf{Checkpoints without an official four-suite release.} GR00T N1.7 has no released checkpoint that covers the four suites. We fine-tune the released base model with the official fine-tuning script on the four official LIBERO datasets with no-op actions removed, which hold 1{,}693 demonstrations and 273{,}465 frames. The language model and the vision encoder stay frozen, and the projector, the diffusion transformer and the vision-language layer norm are trained. Training runs for 80{,}000 steps on two GPUs with a global batch size of 640, AdamW, a peak learning rate of $10^{-4}$ with a cosine schedule and a warm-up ratio of 0.05, a weight decay of $10^{-5}$, a state dropout of 0.2 and colour jitter. OTTER is trained with the authors' code on the same four datasets, sampled in proportion to their number of frames. It uses a frozen CLIP ViT-L/14, a chunk of 12 actions, a batch size of 64, a peak learning rate of $3\times10^{-4}$ that decays to zero after 2{,}000 warm-up steps, a weight decay of 0.01 and 80{,}000 steps on one RTX A6000. The GuidedVLA checkpoint is the one released by its authors for their full configuration with object, depth and skill heads. According to the released configuration it starts from the $\pi_0$ base model and is trained for 30{,}000 steps with a batch size of 64 on the LIBERO dataset released by the authors, which holds the 40 tasks of the four suites in 1{,}722 demonstrations and 277{,}947 frames with object masks and skill labels.

\textbf{ReGuide in simulation.} The three wrapped backbones share demonstration references, intervention criteria and the execution protocol. Only the chunk length and the number of executed actions differ, which are 10 and 5 for $\pi_{0.5}$, 16 and 8 for GR00T N1.7, and 8 and 8 for OpenVLA-OFT. The grounding module is replaced by the simulator. The identity of the instructed object and destination is read from the specification of the cell, and the poses and extents of the entities are read from the simulator state. The instruction text is used only by the policy and by the retrieval of habitual referents. Appendix~\ref{app:perception_grounding} evaluates perception-based grounding in simulation, and Appendix~\ref{app:real} describes the detector used on the real robot.

\section{Additional Simulation Results}
\label{app:results}

\subsection{Pre-contact Initialization Diagnostic}
\label{app:precontact_diag}

\textbf{Protocol.}
We isolate configuration transfer by testing whether a demonstration-derived starting pose restores local grasping under recomposition. Using $\pi_{0.5}$, we compare the default start with a single transport to the instructed object's demonstration-derived pre-contact target at the first decision. After arrival, the pending action chunk is discarded and the policy is queried on the true observation, with all subsequent ReGuide interventions and action constraints disabled. Both conditions use the same 50 initial states per cell and the same episode budget.

The diagnostic covers 10 original tasks and 8 recomposed cells from the Target Object and Task Scene axes, totaling 1800 episodes. The pooled original-task analysis uses the 8 single-clause tasks. O5 and O6 are reported separately because the bare-condition instrumentation tracks only the first clause's object, making their grasp measurements asymmetric across conditions. This adjustment was recorded before the affected cells were run. All 900 initialization transports arrived, with no gripper--object contact or object displacement before arrival.

\begin{figure}[htbp]
    \centering
    \includegraphics[width=\linewidth]{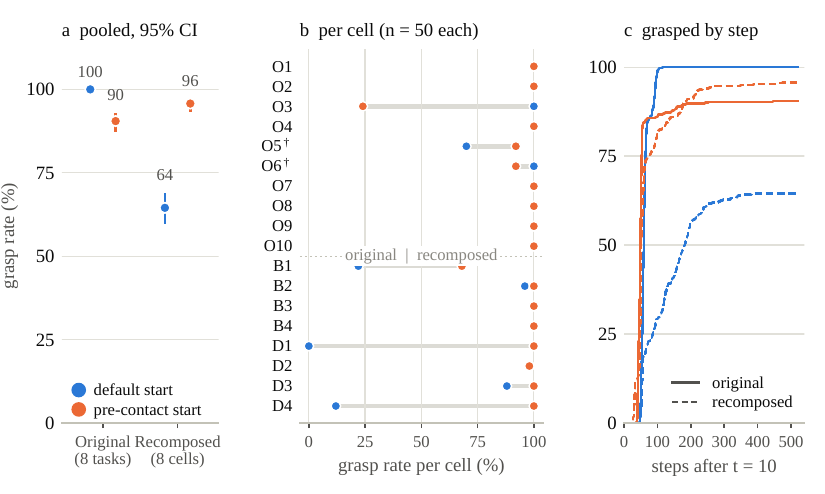}
    \caption{\textbf{Pre-contact initialization diagnostic.}
    Default and pre-contact starts for frozen $\pi_{0.5}$.
    (a) Pooled grasp success with 95\% confidence intervals.
    (b) Per-cell grasp success; O5 and O6 are multi-clause tasks excluded from the pooled original-task results.
    (c) Cumulative grasp success from the first decision, including initialization time.}
    \label{fig:precontact_diag}
\end{figure}

\begin{table}[htbp]
\centering
\caption{\textbf{Diagnostic outcomes.}
Grasp requires lifting the instructed object more than 3\,cm with attachment maintained. Task success uses the original benchmark criterion. All trials in the pooled cells are retained.}
\label{tab:precontact_diag}
\small
\begin{tabular}{llcc}
\toprule
\textbf{Metric} & \textbf{Task type}
& \textbf{Default start} & \textbf{Pre-contact start} \\
\midrule
Grasp & Original   & 400/400 & 362/400 \\
      & Recomposed & 258/400 & 383/400 \\
\midrule
Task success & Original   & 397/400 & 343/400 \\
             & Recomposed & 136/400 & 266/400 \\
\bottomrule
\end{tabular}
\end{table}

\textbf{Recovery of local grasping.}
On recomposed cells, grasp success increases from 64.5\% to 95.8\%. Of the paired trials, 131 succeed only with pre-contact initialization and 6 only with the default start (exact McNemar $p<10^{-31}$). The median time from policy takeover to grasp decreases from 116 to 35 steps. Recovery is especially pronounced on D1 and D4, where grasp success rises from 0\% and 12\%, respectively, to 100\%. On D1, task success remains 0/50 despite recovered grasping, locating the remaining difficulty in the carry stage.

\subsection{Per-Cell Results}

\begin{figure}[htbp]
\centering
\includegraphics[width=\linewidth]{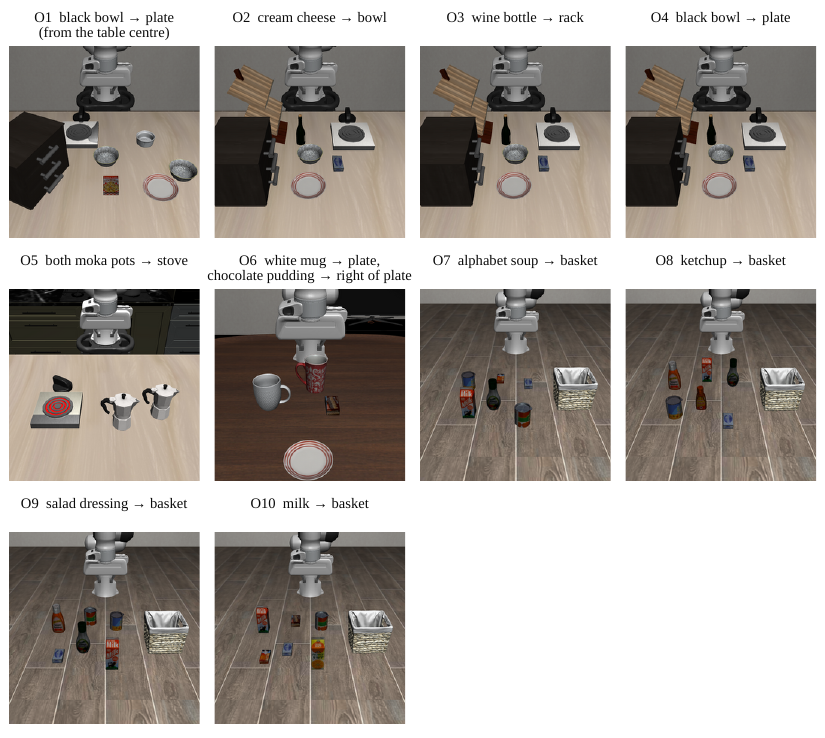}
\caption{The ten original tasks O1 to O10 of Table~\ref{tab:orig}, each shown at the first initial state of its evaluation bank.}
\label{fig:orig_tasks}
\end{figure}

\begin{table}[htbp]
\centering
\caption{Successes out of 50 per composition cell.}
\label{tab:percell}
\resizebox{\linewidth}{!}{%
\begin{tabular}{lccccccccccccccccc}
\toprule
\textbf{Method} & A1 & A2 & A3 & A4 & B1 & B2 & B3 & B4 & C1 & C2 & C3 & C4 & D1 & D2 & D3 & D4 & $\Sigma$ \\
\midrule
OpenVLA-OFT & 0 & 0 & 27 & 0 & 3 & 0 & 0 & 0 & 3 & 0 & 31 & 35 & 0 & 13 & 0 & 0 & 112 \\
$\pi_0$ & 0 & 0 & 10 & 0 & 0 & 0 & 3 & 0 & 0 & 7 & 0 & 47 & 0 & 0 & 1 & 0 & 68 \\
$\pi_{0.5}$ & 1 & 5 & 42 & 33 & 4 & 13 & 9 & 24 & 3 & 19 & 27 & 50 & 0 & 42 & 37 & 3 & 312 \\
X-VLA & 0 & 47 & 31 & 0 & 1 & 6 & 12 & 0 & 0 & 2 & 16 & 50 & 0 & 0 & 0 & 0 & 165 \\
GR00T N1.7 & 0 & 0 & 8 & 21 & 2 & 1 & 10 & 6 & 0 & 1 & 9 & 46 & 0 & 0 & 0 & 0 & 104 \\
\midrule
OTTER & 0 & 0 & 0 & 0 & 0 & 0 & 0 & 0 & 0 & 0 & 15 & 11 & 0 & 0 & 1 & 0 & 27 \\
GuidedVLA & 0 & 0 & 4 & 0 & 0 & 0 & 13 & 0 & 0 & 1 & 2 & 50 & 0 & 0 & 1 & 0 & 71 \\
APT & 0 & 31 & 5 & 38 & 3 & 1 & 49 & 14 & 0 & 0 & 1 & 50 & 0 & 3 & 0 & 0 & 195 \\
CAG ($\pi_{0.5}$) & 2 & 32 & 39 & 33 & 8 & 13 & 25 & 24 & 5 & 6 & 29 & 48 & 0 & 37 & 30 & 0 & 331 \\
\midrule
ReGuide (OpenVLA-OFT) & 50 & 30 & 47 & 47 & 2 & 49 & 24 & 23 & 30 & 21 & 50 & 44 & 49 & 34 & 40 & 26 & 566 \\
ReGuide ($\pi_{0.5}$) & 50 & 48 & 46 & 48 & 49 & 49 & 49 & 39 & 48 & 46 & 46 & 49 & 49 & 42 & 48 & 50 & 756 \\
ReGuide (GR00T N1.7) & 50 & 27 & 25 & 36 & 2 & 23 & 48 & 32 & 32 & 2 & 48 & 36 & 49 & 43 & 50 & 50 & 553 \\
\bottomrule
\end{tabular}
}
\end{table}

\begin{table}[htbp]
\centering
\caption{Successes out of 50 per original task. O1 black bowl from table centre to plate, O2 cream cheese to bowl, O3 wine bottle to rack, O4 bowl to plate, O5 both moka pots to stove, O6 white mug to plate and chocolate pudding to the right of the plate, O7 alphabet soup, O8 ketchup, O9 salad dressing, O10 milk to basket.}
\label{tab:orig}
\resizebox{0.85\linewidth}{!}{%
\begin{tabular}{lccccccccccc}
\toprule
\textbf{Method} & O1 & O2 & O3 & O4 & O5 & O6 & O7 & O8 & O9 & O10 & $\Sigma$ \\
\midrule
OpenVLA-OFT & 49 & 49 & 50 & 50 & 46 & 46 & 50 & 50 & 48 & 50 & 488 \\
$\pi_0$ & 49 & 49 & 40 & 50 & 5 & 40 & 49 & 48 & 50 & 50 & 430 \\
$\pi_{0.5}$ & 50 & 50 & 47 & 50 & 31 & 44 & 50 & 50 & 48 & 50 & 470 \\
X-VLA & 49 & 49 & 46 & 50 & 48 & 47 & 50 & 50 & 50 & 49 & 488 \\
GR00T N1.7 & 50 & 49 & 49 & 49 & 34 & 44 & 50 & 50 & 48 & 50 & 473 \\
\midrule
OTTER & 46 & 42 & 31 & 50 & 21 & 30 & 41 & 50 & 50 & 50 & 411 \\
GuidedVLA & 49 & 49 & 43 & 49 & 39 & 45 & 50 & 50 & 50 & 50 & 474 \\
APT & 50 & 50 & 48 & 50 & 14 & 45 & 50 & 50 & 50 & 50 & 457 \\
CAG ($\pi_{0.5}$) & 49 & 46 & 44 & 50 & 10 & 17 & 48 & 47 & 46 & 50 & 407 \\
\midrule
ReGuide (OpenVLA-OFT) & 49 & 47 & 50 & 50 & 47 & 42 & 50 & 50 & 49 & 50 & 484 \\
ReGuide ($\pi_{0.5}$) & 49 & 50 & 50 & 49 & 34 & 40 & 49 & 50 & 50 & 50 & 471 \\
ReGuide (GR00T N1.7) & 49 & 49 & 50 & 49 & 33 & 46 & 49 & 47 & 49 & 50 & 471 \\
\bottomrule
\end{tabular}
}
\end{table}

\begin{table}[htbp]
\centering
\caption{ReGuide against its bare backbone on paired trials. Wins are trials solved only by ReGuide, losses trials solved only by the bare policy, and $p$ is the exact McNemar test.}
\label{tab:paired}
\small
\fitwidth{%
\begin{tabular}{llcccccc}
\toprule
\textbf{Backbone} & \textbf{Split} & \textbf{Bare} & \textbf{ReGuide} & $\boldsymbol{\Delta}$ (pp) & \textbf{Wins} & \textbf{Losses} & $p$ \\
\midrule
OpenVLA-OFT & Target Region & 27/200 & 174/200 & +73.5 & 148 & 1 & $<$0.001 \\
 & Target Object & 3/200 & 98/200 & +47.5 & 97 & 2 & $<$0.001 \\
 & Unseen Object & 69/200 & 145/200 & +38.0 & 81 & 5 & $<$0.001 \\
 & Task Scene & 13/200 & 149/200 & +68.0 & 136 & 0 & $<$0.001 \\
 & Comp. & 112/800 & 566/800 & +56.8 & 462 & 8 & $<$0.001 \\
 & Orig. & 488/500 & 484/500 & -0.8 & 6 & 10 & 0.454 \\
\midrule
$\pi_{0.5}$ & Target Region & 81/200 & 192/200 & +55.5 & 116 & 5 & $<$0.001 \\
 & Target Object & 50/200 & 186/200 & +68.0 & 139 & 3 & $<$0.001 \\
 & Unseen Object & 99/200 & 189/200 & +45.0 & 95 & 5 & $<$0.001 \\
 & Task Scene & 82/200 & 189/200 & +53.5 & 115 & 8 & $<$0.001 \\
 & Comp. & 312/800 & 756/800 & +55.5 & 465 & 21 & $<$0.001 \\
 & Orig. & 470/500 & 471/500 & +0.2 & 22 & 21 & 1.000 \\
\midrule
GR00T N1.7 & Target Region & 29/200 & 138/200 & +54.5 & 114 & 5 & $<$0.001 \\
 & Target Object & 19/200 & 105/200 & +43.0 & 92 & 6 & $<$0.001 \\
 & Unseen Object & 56/200 & 118/200 & +31.0 & 76 & 14 & $<$0.001 \\
 & Task Scene & 0/200 & 192/200 & +96.0 & 192 & 0 & $<$0.001 \\
 & Comp. & 104/800 & 553/800 & +56.1 & 474 & 25 & $<$0.001 \\
 & Orig. & 473/500 & 471/500 & -0.4 & 18 & 20 & 0.871 \\
\bottomrule
\end{tabular}%
}
\end{table}

\begin{table}[htbp]
\centering
\caption{Failures of ReGuide on the 800 composition trials, by the stage at which the episode was lost.}
\label{tab:residual}
\small
\fitwidth{%
\begin{tabular}{lccccl}
\toprule
\textbf{ReGuide on} & \textbf{Failures} & \textbf{Grasp} & \textbf{Carry} & \textbf{After arrival} & \textbf{Cells with most failures} \\
\midrule
$\pi_{0.5}$ & 44 & 11 & 3 & 30 & B4 (11), D2 (8) \\
GR00T N1.7 & 247 & 106 & 36 & 105 & B1 (48), C2 (48) \\
OpenVLA-OFT & 234 & 113 & 32 & 89 & B1 (48), C2 (29) \\
\bottomrule
\end{tabular}%
}
\end{table}

Tables~\ref{tab:percell} and~\ref{tab:orig} give the successes out of 50 for every cell and every original task (Fig.~\ref{fig:orig_tasks}), and Table~\ref{tab:paired} the paired tests of ReGuide against its bare backbone. Without ReGuide the successes concentrate on the seven cells of the Goal table scene, A2 to A4, B3, B4, C2 and C4, which hold 187 of the 195 composition successes of APT and 92 of the 104 of GR00T N1.7, and within them on C4, where the trained behaviour already satisfies the instruction.

\textbf{Remaining failures.} Table~\ref{tab:residual} splits the failures of ReGuide by the last stage that was reached. For GR00T N1.7 and OpenVLA-OFT the failures concentrate in cells where the instructed object is reached and the resumed policy does not complete the grasp, as in B1, where the object is held in 8 and 7 of 50 episodes against 50 for $\pi_{0.5}$.

\subsection{Intervention Statistics}
\label{app:intervention_stats}

\begin{figure}[htbp]
\centering
\includegraphics[width=\linewidth]{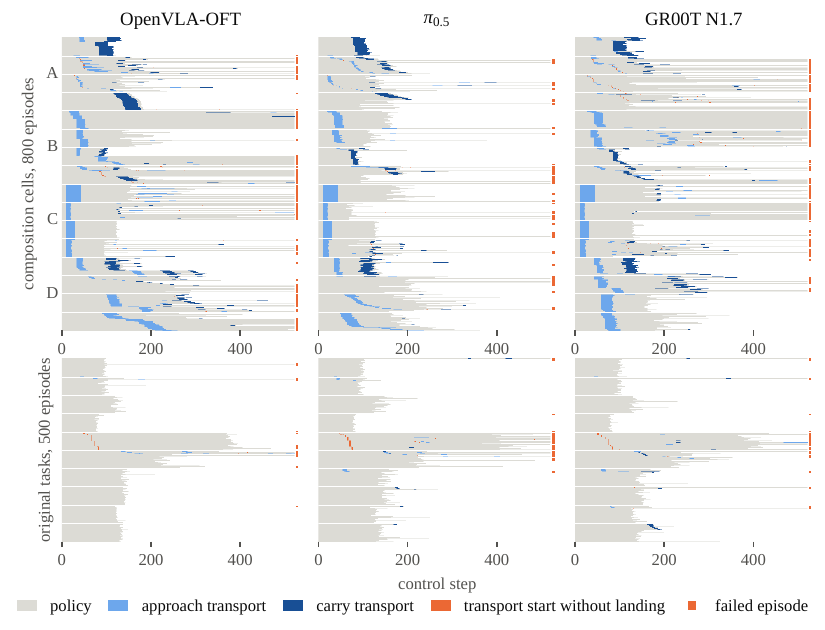}
\caption{Every ReGuide episode in simulation as one row, composition cells above and original tasks below, sorted within each cell by the start of the first transport. Light blue marks the steps of an approach transport and dark blue those of a carry transport. Orange marks a transport start without a landing, that is a transport that was dissolved, hijacked or cancelled, and the squares on the right mark failed episodes.}
\label{fig:raster}
\end{figure}

\begin{figure}[htbp]
\centering
\includegraphics[width=\linewidth]{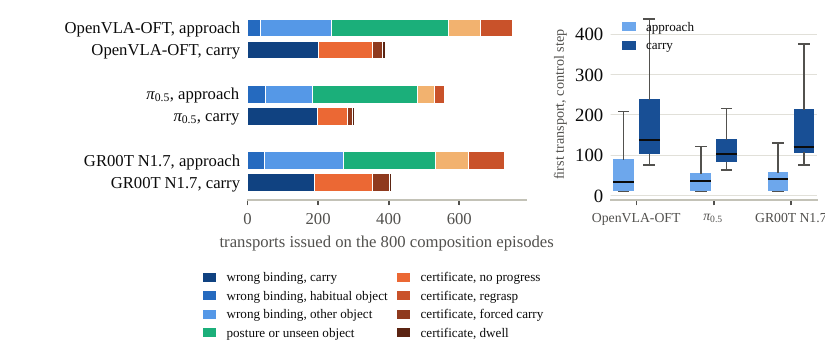}
\caption{Causes of the transports issued on the 800 composition episodes (left) and the step of the first transport of an episode (right, boxes are quartiles and whiskers the range without outliers). Wrong binding denotes a predicted commitment to an incorrect object during approach or a habitual destination during carrying. Posture covers the posture mismatch and the unseen-object rule, and the remaining causes are the failure to bind and the completion certificates.}
\label{fig:causes}
\end{figure}

\begin{figure}[htbp]
\centering
\includegraphics[width=\linewidth]{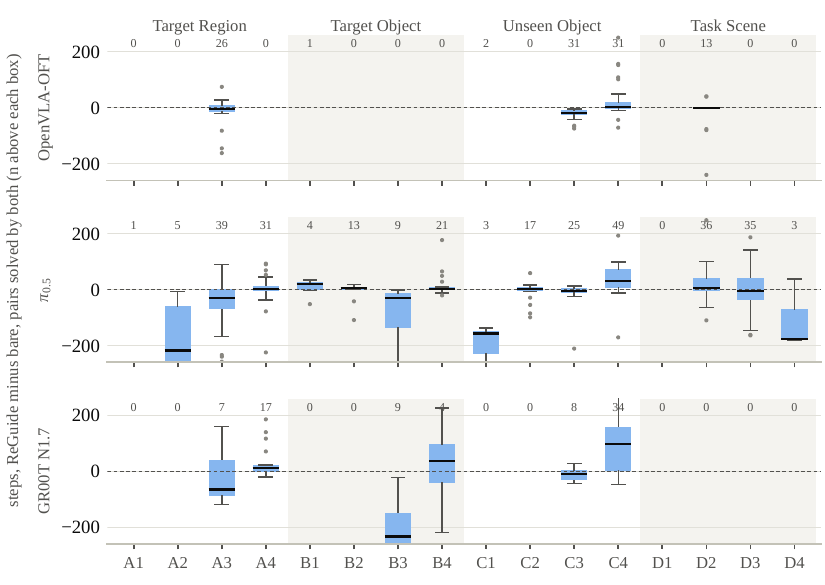}
\caption{Change of the episode length under ReGuide per composition cell, in steps, over the pairs of trials solved by both the bare backbone and ReGuide. The number of pairs is given above each box, and cells with fewer than three pairs are left empty.}
\label{fig:steps_paired}
\end{figure}

An episode is assigned to one class with the priority wrong binding, posture, reactive, none. The posture class contains the unseen-object rule, which by construction fires at the first decision of every episode of the Unseen Object axis. The reactive class contains stalled-approach recovery and completion certificates. These classes distinguish referent mismatch, configuration compatibility and interaction progress. The classes are the verdicts of the wrapper and not ground-truth labels of the binding. On the original task with two clauses in Kitchen 8, a transport that is issued at the switch of the clause is cancelled one step later in every episode, and these one-step transports are not counted. With them the transport rates on original tasks are 20.4, 24.2 and 14.4\%. The length change compares trials that both methods solve from the same initial state, which are two independent rollouts.

Wrong-binding corrections occur in 1/500, 2/500 and 0/500 original episodes for $\pi_{0.5}$, GR00T N1.7 and OpenVLA-OFT, respectively. Among successful composition episodes with at least one transport, the median fractions of control steps spent in transport are 14.3\%, 14.6\% and 15.4\%, respectively.

Figs.~\ref{fig:raster}--\ref{fig:steps_paired} detail transport timing, causes, and paired episode lengths. On the 800 composition episodes $\pi_{0.5}$ issues 559 approach and 304 carry transports, GR00T N1.7 729 and 408, and OpenVLA-OFT 751 and 391. The approach transports mostly start in the first hundred steps, with a median first transport at step 35, 42 and 34 for $\pi_{0.5}$, GR00T N1.7 and OpenVLA-OFT, and the carry transports follow the grasp. On the original tasks transports are rare, and the short marks without a landing lie almost entirely on the two-clause tasks O5 and O6, 56 of 56 for $\pi_{0.5}$, 56 of 62 for GR00T N1.7 and 55 of 55 for OpenVLA-OFT.

\subsection{Ablation Details}
\label{app:ablation}

\begin{table}[htbp]
\centering
\caption{Ablation on $\pi_{0.5}$, successes out of 50 per cell.}
\label{tab:ablcells}
\resizebox{\linewidth}{!}{%
\begin{tabular}{lccccccccccccccccc}
\toprule
\textbf{Variant} & A1 & A2 & A3 & A4 & B1 & B2 & B3 & B4 & C1 & C2 & C3 & C4 & D1 & D2 & D3 & D4 & $\Sigma$ \\
\midrule
Naive transport & 35 & 2 & 0 & 50 & 7 & 50 & 50 & 34 & 16 & 38 & 48 & 49 & 39 & 34 & 46 & 50 & 548 \\
w/o competing referents & 34 & 49 & 45 & 45 & 49 & 49 & 14 & 42 & 44 & 48 & 44 & 47 & 38 & 43 & 48 & 48 & 687 \\
w/o chunk commitment & 49 & 41 & 43 & 45 & 45 & 49 & 50 & 23 & 25 & 47 & 49 & 42 & 44 & 44 & 49 & 46 & 691 \\
w/o pre-contact sets & 50 & 48 & 43 & 33 & 28 & 49 & 49 & 15 & 23 & 29 & 50 & 49 & 47 & 40 & 29 & 25 & 607 \\
w/o bounded transport & 50 & 31 & 41 & 46 & 45 & 41 & 49 & 40 & 49 & 50 & 38 & 43 & 41 & 44 & 47 & 46 & 701 \\
ReGuide (full) & 50 & 48 & 46 & 48 & 49 & 49 & 49 & 39 & 48 & 46 & 46 & 49 & 49 & 42 & 48 & 50 & 756 \\
\bottomrule
\end{tabular}
}
\end{table}

\begin{table}[htbp]
\centering
\caption{Stage counts of the ablation over 800 trials. Object and Destination count episodes in which the end-effector reached the instructed object and the held object reached the instructed destination. Successful trials without a logged stage event are excluded from failure counts, so these need not equal differences between adjacent stage counts.}
\label{tab:ablstage}
\small
\fitwidth{%
\begin{tabular}{lccccccc}
\toprule
 & \multicolumn{4}{c}{\textbf{Episodes reaching the stage}} & \multicolumn{3}{c}{\textbf{Episodes lost}} \\
\cmidrule(lr){2-5}\cmidrule(lr){6-8}
\textbf{Variant} & Object & Held & Destination & Success & Grasp & Carry & After arrival \\
\midrule
Naive transport & 795 & 708 & 651 & 548 & 92 & 57 & 103 \\
w/o competing referents & 796 & 783 & 737 & 687 & 17 & 46 & 50 \\
w/o chunk commitment & 789 & 758 & 726 & 691 & 42 & 31 & 36 \\
w/o pre-contact sets & 790 & 708 & 674 & 607 & 92 & 32 & 69 \\
w/o bounded transport & 787 & 771 & 759 & 701 & 29 & 12 & 58 \\
ReGuide (full) & 798 & 789 & 785 & 756 & 11 & 3 & 30 \\
\bottomrule
\end{tabular}%
}
\end{table}

\begin{table}[htbp]
\centering
\caption{Sensitivity to the declared constants and to two design forms, on the 800 composition trials with $\pi_{0.5}$. The last columns count trials solved only by the default or only by the variant.}
\label{tab:variants}
\resizebox{\linewidth}{!}{%
\begin{tabular}{lllccccc}
\toprule
\textbf{Quantity} & \textbf{Default} & \textbf{Variant} & $k/800$ & \textbf{Success} (\%) & \textbf{Default only} & \textbf{Variant only} & $p$ \\
\midrule
Default & -- & -- & 756 & 94.5$_{\pm1.6}$ & -- & -- & -- \\
\midrule
$n$ & 3 & 1 & 760 & 95.0$_{\pm1.5}$ & 27 & 31 & 0.694 \\
$n$ & 3 & 5 & 759 & 94.9$_{\pm1.5}$ & 25 & 28 & 0.784 \\
$\varepsilon$ (mm) & 4 & 2 & 755 & 94.4$_{\pm1.6}$ & 34 & 33 & 1.000 \\
$\varepsilon$ (mm) & 4 & 8 & 752 & 94.0$_{\pm1.7}$ & 36 & 32 & 0.716 \\
trigger level & $Q_{.25}$ & $Q_{.50}$ & 766 & 95.8$_{\pm1.4}$ & 25 & 35 & 0.245 \\
upper level & $Q_{.95}$ & $Q_{.75}$ & 751 & 93.9$_{\pm1.7}$ & 36 & 31 & 0.625 \\
lower level & $Q_{.05}$ & $Q_{.25}$ & 760 & 95.0$_{\pm1.5}$ & 32 & 36 & 0.716 \\
\midrule
descent tolerance $r_e$ & min of scatter and half-extent & max of the two & 754 & 94.2$_{\pm1.6}$ & 39 & 37 & 0.909 \\
acceptance region $\Omega_d$ & range $\cap$ scatter & scatter only & 750 & 93.8$_{\pm1.7}$ & 38 & 32 & 0.550 \\
\bottomrule
\end{tabular}
}
\end{table}

\textbf{Component groups.} Table~\ref{tab:ablcells} retains the variant names used in Table~\ref{tab:ablation_mech}. The first two removals concern referent monitoring in Sec.~\ref{sec:semantic}, and the last two concern configuration transfer and transport in Sec.~\ref{sec:geometric}.

\textit{Naive transport} uses the same grounding information, servoing once to 0.25\,m above the instructed object at the start and once to 0.25\,m above the destination after the object is held, then handing back. Without \textit{competing referents}, no habitual or watch set is compiled and objects are identified by name. Without \textit{chunk commitment}, proximity replaces the endpoint test, and its accompanying mechanisms are removed, including suppression on the instructed object, carry gates, first-decision guidance for unseen objects, transport dissolution, failure-to-bind monitoring and completion certificates. Without \textit{pre-contact sets}, a fixed 0.25\,m target replaces the interaction configuration, with no posture target, descent tolerance or acceptance region. Without \textit{bounded transport}, commands use clipped residuals without step bounds or tracking gains, and omit the height profile and floor. Table~\ref{tab:ablstage} locates the resulting losses within the interaction stages shown in Fig.~\ref{fig:ablation_fig}.

\textbf{Constants and design forms.} Table~\ref{tab:variants} changes one declared constant or quantile level at a time, and two design forms. No variant differs significantly from the default under the paired McNemar test.

\textbf{Targeted controls.}
Targeted controls separate local policy execution and post-hand-back coordination. Each control uses the frozen $\pi_{0.5}$ checkpoint, the same demonstration statistics, and the 800 paired composition initial states under the protocol of Appendix~\ref{app:bench}. Full ReGuide results are from the main evaluation. Table~\ref{tab:targeted_cells} gives per-cell results. Action replacement is described in Appendix~\ref{app:handback_replacement}.

\textit{Without hand-back assistance.}
After the first transport lands, we disable the descent corridor's lateral correction, the carry-height floor, completion-certificate recovery (rearming, regrasping, and hover-stall escort), immediate escort after landing, and escort descent and release. Failure-to-bind triggering, subsequent approach and carry transports, their targets, step laws, and budgets remain active. The switch occurs in 656/800 episodes. Success falls from 756/800 to 677/800, a paired difference of $-9.9$ percentage points with 95\% bootstrap interval $[-12.8,-7.0]$. Full alone succeeds on 113 pairs and this variant alone on 34 ($p=4.2\times10^{-11}$). Grasp success remains 97.1\% versus 98.6\% for full, while destination arrival falls from 98.1\% to 93.9\%. Most additional failures occur after grasping (Table~\ref{tab:handback_stages}).

\begin{table}[htbp]
\centering
\caption{\textbf{Post-hand-back assistance by composition cell.} Successes out of 50. \textit{No assistance} disables post-hand-back constraints and completion assistance.}
\label{tab:targeted_cells}
\begin{tabular}{lrr}
\toprule
\textbf{Cell} & \textbf{Full} & \textbf{No assistance} \\
\midrule
A1 & 50 & 48 \\
A2 & 48 & 49 \\
A3 & 46 & 43 \\
A4 & 48 & 48 \\
B1 & 49 & 35 \\
B2 & 49 & 48 \\
B3 & 49 & 50 \\
B4 & 39 & 20 \\
C1 & 48 & 45 \\
C2 & 46 & 43 \\
C3 & 46 & 49 \\
C4 & 49 & 36 \\
D1 & 49 & 36 \\
D2 & 42 & 46 \\
D3 & 48 & 44 \\
D4 & 50 & 37 \\
\midrule
Total & 756 & 677 \\
\bottomrule
\end{tabular}
\end{table}

\subsection{Guidance Weight of CAG}

\begin{table}[htbp]
\centering
\caption{CAG on $\pi_{0.5}$ under two guidance weights (success rate, \%).}
\label{tab:cag}
\small
\setlength{\tabcolsep}{4pt}
\fitwidth{%
\begin{tabular}{lcccccc}
\toprule
\textbf{Method} & \textbf{Orig.} & \textbf{Target Region} & \textbf{Target Object} & \textbf{Unseen Object} & \textbf{Task Scene} & \textbf{Comp.\ Avg.} \\
\midrule
Bare $\pi_{0.5}$ & 94.0 & 40.5 & 25.0 & 49.5 & 41.0 & 39.0 \\
CAG, $\omega=1.5$ & 81.4 & 53.0 & 35.0 & 44.0 & 33.5 & 41.4 \\
CAG, $\omega=2.0$ & 64.6 & 59.0 & 27.0 & 38.0 & 20.0 & 36.0 \\
\bottomrule
\end{tabular}%
}
\end{table}

Table~\ref{tab:cag} compares the guidance weight 1.5 used in Table~\ref{tab:composition_main} with 2.0, the example value in the authors' repository.

\subsection{Perception-Based Grounding}
\label{app:perception_grounding}

\textbf{Setup.}
We replace oracle grounding with an RGB-D perception module for ReGuide on $\pi_{0.5}$, keeping referent monitoring, geometric execution, demonstration statistics, and the execution budget unchanged. Inputs are the instruction, a 512-pixel agentview RGB-D image, and declared category geometry from the object assets, analogous to the measured dimensions used on the real robot. GroundingDINO tiny uses box and text thresholds of 0.25 with the declared scene category vocabulary. Detected point-cloud segments are assigned to declared instances by maximum-weight one-to-one matching, with geometric gates on resting top height and footprint dimensions. Spatial phrases in the instruction distinguish instances of the same category. Depth back-projection provides positions, using the center of the declared top-surface slice for objects and the point-cloud bounding-box center for fixtures.

Registration occurs once when the instruction is issued, without subsequent visual tracking. The grounding module reads neither simulator entity identities nor ground-truth poses. Episodes in which the instructed object or destination cannot be registered count as failures, with no oracle fallback.

\begin{table}[htbp]
\centering
\caption{\textbf{ReGuide with oracle and perception-based grounding on $\pi_{0.5}$ (compositional success rate, \%).} Each composition axis contains 200 trials. Subscripts are 95\% Wilson confidence half-widths. Registration failures are included.}
\label{tab:perception_grounding}
\small
\resizebox{\linewidth}{!}{%
\begin{tabular}{lccccc}
\toprule
\textbf{Grounding} & \textbf{Target Region} & \textbf{Target Object} & \textbf{Unseen Object} & \textbf{Task Scene} & \textbf{Comp.\ Avg.} \\
\midrule
Oracle & $96.0_{\pm2.8}$ & $93.0_{\pm3.6}$ & $94.5_{\pm3.2}$ & $94.5_{\pm3.2}$ & $94.5_{\pm1.6}$ \\
Perception & $86.0_{\pm4.8}$ & $86.5_{\pm4.7}$ & $91.5_{\pm3.9}$ & $86.0_{\pm4.8}$ & $87.5_{\pm2.3}$ \\
\bottomrule
\end{tabular}%
}
\end{table}

\textbf{Results.}
Compositional success remains 87.5\% (700/800), compared with 94.5\% (756/800) under oracle grounding, retaining a 48.5 percentage-point gain over the bare policy. The paired difference to oracle is $-7.0$ percentage points, with a 95\% paired-bootstrap interval of $[-9.8,-4.4]$. Perception alone succeeds on 32 pairs and oracle alone on 88.

\textbf{Error analysis.}
Identity confusion affects referent matching, while fragmented fixture point clouds and stale estimates after object motion affect the transferred configurations.

\FloatBarrier
\subsection{Hand-Back Action Replacement}
\label{app:handback_replacement}

\textbf{Protocol.}
We test local policy execution while retaining ReGuide's referent monitoring, configuration transfer and execution support. From the first transport landing until episode termination, the action selected from the policy chunk is replaced before applying the wrapper's constraints. The VLA is still queried on the true observation at the standard cadence, and its predicted chunks continue to feed the commitment detector. Subsequent transports, motion envelopes and completion certificates remain active. Episodes without a hand-back follow the original execution path. Each condition uses the same 16 composition cells and 50 initial states per cell as full ReGuide.

\textbf{Replacement actions.}
The \textit{default action} holds the hand-back horizontal position and orientation, descends with the demonstrated step size to the demonstrated median grasp height, then closes and holds the gripper. Following a post-grasp hand-back, it instead descends to the demonstrated release height while keeping the gripper closed. In the 44 episodes without an available demonstration target, it closes and holds at the landing height. The \textit{hold} condition sets translation and rotation increments to zero and retains the gripper command at the switch. Both replacements remain subject to the unchanged wrapper. Replacement covers 73.7\% and 77.6\% of execution steps for default and hold, respectively.

\textbf{Results.}
Full ReGuide solves 756/800 trials (94.5\%, 95\% Wilson interval $[92.7,95.9]$), compared with 305/800 for default (38.1\%, $[34.8,41.5]$) and 224/800 for hold (28.0\%, $[25.0,31.2]$). Full exceeds default by 56.4 percentage points with paired-bootstrap interval $[52.6,60.1]$ (470 full-only and 19 default-only successes, $p=9.4\times10^{-114}$), and hold by 66.5 points with interval $[63.0,70.0]$ (547 full-only and 15 hold-only successes, $p=1.5\times10^{-140}$). Grasp success is 98.6\%, 44.0\%, and 33.5\%, and destination arrival is 98.1\%, 42.1\%, and 32.2\%, respectively. Table~\ref{tab:handback_cells} gives the per-cell counts.

\begin{table}[htbp]
\centering
\caption{\textbf{Outcomes by first hand-back stage.} Successes/trials with percentages in parentheses. Groups are defined within each rollout, so their membership can differ across conditions.}
\label{tab:handback_split}
\begin{tabular}{lccc}
\toprule
\textbf{Action} & \textbf{No hand-back} & \textbf{Before grasp} & \textbf{After grasp} \\
\midrule
VLA (full) & 120/139 (86.3) & 510/531 (96.0) & 126/130 (96.9) \\
Default & 126/147 (85.7) & 52/513 (10.1) & 127/140 (90.7) \\
Hold & 121/141 (85.8) & 0/529 (0.0) & 103/130 (79.2) \\
\bottomrule
\end{tabular}
\end{table}

\textbf{Role of local execution.}
The largest loss occurs when replacement begins before grasping (Table~\ref{tab:handback_split}). Default and hold incur 440 and 521 post-hand-back grasp failures, versus 6 for full (Table~\ref{tab:handback_stages}). Replacement after grasping is less disruptive, consistent with continued transport and release assistance during placement. Completion escort performs the final release in 86 full episodes (83 successful), 128 default episodes (119 successful), and 63 hold episodes (all successful). Together with the initialization diagnostic, these controls link access to an interaction configuration with the value of executing the local policy from it.

\begin{table}[htbp]
\centering
\caption{\textbf{Failure stages after hand-back.} Counts over 800 trials per condition. Failure categories after \textit{No hand-back} apply only to episodes with a hand-back. Each row sums to 800. \textit{No assistance} is the targeted control in Appendix~\ref{app:ablation}.}
\label{tab:handback_stages}
\small
\fitwidth{%
\begin{tabular}{lrrrrrr}
\toprule
\textbf{Condition} & \textbf{Success} & \shortstack{\textbf{No hand-back}\\\textbf{failure}} & \shortstack{\textbf{No}\\\textbf{grasp}} & \shortstack{\textbf{No destination}\\\textbf{arrival}} & \shortstack{\textbf{No}\\\textbf{release}} & \shortstack{\textbf{Failure after}\\\textbf{release}} \\
\midrule
Full & 756 & 19 & 6 & 1 & 1 & 17 \\
Default & 305 & 21 & 440 & 11 & 2 & 21 \\
Hold & 224 & 20 & 521 & 7 & 5 & 23 \\
No assistance & 677 & 17 & 16 & 23 & 17 & 50 \\
\bottomrule
\end{tabular}%
}
\end{table}

\begin{table}[htbp]
\centering
\caption{\textbf{Hand-back action replacement by cell.} Successes out of 50.}
\label{tab:handback_cells}
\begin{tabular}{lrrr@{\qquad}lrrr}
\toprule
\textbf{Cell} & \textbf{Full} & \textbf{Default} & \textbf{Hold} & \textbf{Cell} & \textbf{Full} & \textbf{Default} & \textbf{Hold} \\
\midrule
A1 & 50 & 50 & 49 & C1 & 48 & 0 & 0 \\
A2 & 48 & 2 & 1 & C2 & 46 & 1 & 0 \\
A3 & 46 & 9 & 9 & C3 & 46 & 0 & 0 \\
A4 & 48 & 47 & 42 & C4 & 49 & 0 & 0 \\
B1 & 49 & 0 & 0 & D1 & 49 & 34 & 0 \\
B2 & 49 & 15 & 13 & D2 & 42 & 42 & 37 \\
B3 & 49 & 37 & 30 & D3 & 48 & 13 & 6 \\
B4 & 39 & 39 & 34 & D4 & 50 & 16 & 3 \\
\bottomrule
\end{tabular}
\end{table}

\subsection{Agent-Based Guidance and Runtime}
\label{app:harness_runtime}

\textbf{Harness VLA protocol.}
We evaluate the official Harness VLA/RPent implementation~\citep{zhang2026harness}, using its released guides and memory (\texttt{RLinf/RPent-memory}) and the documented Codex planner with GPT-5.5 and at most 100 rounds. Its tools include VLA contact skills, waypoint motion, gripper commands, SAM3 segmentation, and depth back-projection. We use the same \texttt{pi05\_libero} checkpoint, image preprocessing, 8-dimensional state, and 5 executed actions per query as in the main evaluation. We retain the 10 settling steps, 520-step execution budget, 20\,Hz control, and per-step LIBERO success predicate.

To bound API cost, we run the first paired initial state (seed 0) in each of the 26 cells and compare against the matching bare-policy and ReGuide trials from the main evaluation. Harness uses its SAM3 and depth pipeline, while these ReGuide rollouts use oracle grounding. The released Harness memory remains unchanged throughout evaluation. It contains LIBERO-PRO seed-0 recipes and reference trajectories and is consulted in 16/26 episodes. In 2 original tasks, soup-to-basket and milk-to-basket, the retrieved reference has the same object and destination.

\begin{table}[htbp]
\centering
\caption{\textbf{Exploratory comparison with Harness VLA.} Successes/trials on one paired initial state per cell. A--D denote the four composition axes. These counts are separate from the 50-trial-per-cell evaluation.}
\label{tab:harness_comparison}
\begin{tabular}{lccccccc}
\toprule
\textbf{Method} & \textbf{A} & \textbf{B} & \textbf{C} & \textbf{D} & \textbf{Comp.} & \textbf{Orig.} & \textbf{All} \\
\midrule
Bare $\pi_{0.5}$ & 1/4 & 0/4 & 2/4 & 1/4 & 4/16 & 10/10 & 14/26 \\
Harness VLA & 3/4 & 1/4 & 2/4 & 4/4 & 10/16 & 8/10 & 18/26 \\
ReGuide & 4/4 & 4/4 & 3/4 & 4/4 & 15/16 & 8/10 & 23/26 \\
\bottomrule
\end{tabular}
\end{table}

\textbf{Outcomes and execution.}
ReGuide alone succeeds on 6 pairs and Harness alone on 1 ($p=0.125$). Harness alone succeeds on 7 pairs and the bare policy alone on 3 ($p=0.34$). With one rollout per cell, these are exploratory paired outcomes rather than precise success-rate estimates. In the 17 jointly successful episodes, median steps to success are 279 for Harness and 141 for ReGuide.

\textbf{Planner waiting time and cost.}
Harness has a median episode wall time of 184\,s, including 156\,s spent waiting for the planner. A planner response takes a median of 6.1\,s, a 90th percentile of 12.5\,s, and a maximum of 37.4\,s, with a median of 20 waits per episode. API cost averages \$1.84 per episode.

\textbf{ReGuide timing protocol.}
We measure the runtime of ReGuide with oracle grounding and the bare policy on the same 26 seed-0 initial states. Both use one dedicated RTX A6000 policy server and a Xeon w5-3435X CPU, with no other jobs or video recording. Replanning latency includes the policy query, websocket communication, and client computation, excluding simulator physics and rendering. Timing is collected in separate runs from the success comparison in Table~\ref{tab:harness_comparison}.

\begin{table}[htbp]
\centering
\caption{\textbf{Runtime of oracle-grounding ReGuide and the bare policy.} Timings in milliseconds after the first decision. Each entry reports median / 90th percentile unless otherwise stated.}
\label{tab:reguide_runtime}
\begin{tabular}{lrr}
\toprule
\textbf{Measurement} & \textbf{Bare $\pi_{0.5}$} & \textbf{ReGuide} \\
\midrule
Policy query & 127.2 / 128.1 & 126.7 / 127.7 \\
Replanning latency & 129.0 / 129.9 & 129.2 / 130.3 \\
Client computation per step & 1.60 / 1.81 & 2.14 / 2.61 \\
Client computation (99th percentile) & 2.04 & 3.49 \\
Simulator step (median) & 10.3 & 10.1 \\
\bottomrule
\end{tabular}
\end{table}

\textbf{Runtime results.}
Table~\ref{tab:reguide_runtime} reports steady-state latency. Initialization, including retrieval-encoder loading, takes a median of 2.2\,s versus 28\,ms for the bare policy. These timings use oracle grounding and exclude GroundingDINO. Planner waits and policy replanning are measured per event at their respective frequencies.

\FloatBarrier
\section{Real-World Details}
\label{app:real}

\subsection{Hardware and Control}

The robot is a UFACTORY xArm6 with the xArm gripper. Two Intel RealSense D435i cameras provide aligned RGB-D streams at $640\times480$ and 30\,fps, one fixed at the base and one on the wrist, and the policy receives $480\times480$ crops of both. Control runs at 10\,Hz with absolute joint targets streamed to the controller. A workstation with two RTX 4090 GPUs runs the policy on one GPU and the detector on the other. A safety layer below the wrapper limits the joint step to 0.30\,rad per tick and the joint speed to 2.0\,rad/s, keeps the joints 2$^\circ$ inside the limits of the controller, and enforces a floor of the tool above the table.

\subsection{Policy}

$\pi_{0.5}$ is fully fine-tuned from the base checkpoint with AdamW, gradient clipping at 1.0, a learning rate warmed up over 1600 steps to $5\times10^{-5}$ and then kept constant, an exponential moving average of 0.999, batch size 128 and 16k steps on one H200, which took about 21.5 hours. The final checkpoint is deployed. The observation consists of the two image crops and a 7-dimensional state of six joint angles and the gripper, and the action is a chunk of 10 absolute joint and gripper targets, of which 8 are executed per replan while the next chunk is requested two ticks ahead. Median inference time is 94\,ms with ReGuide and 92\,ms without. Observations are two ticks old when chunk execution begins.

\subsection{Demonstrations}

\begin{figure}[htbp]
\centering
\includegraphics[width=\linewidth]{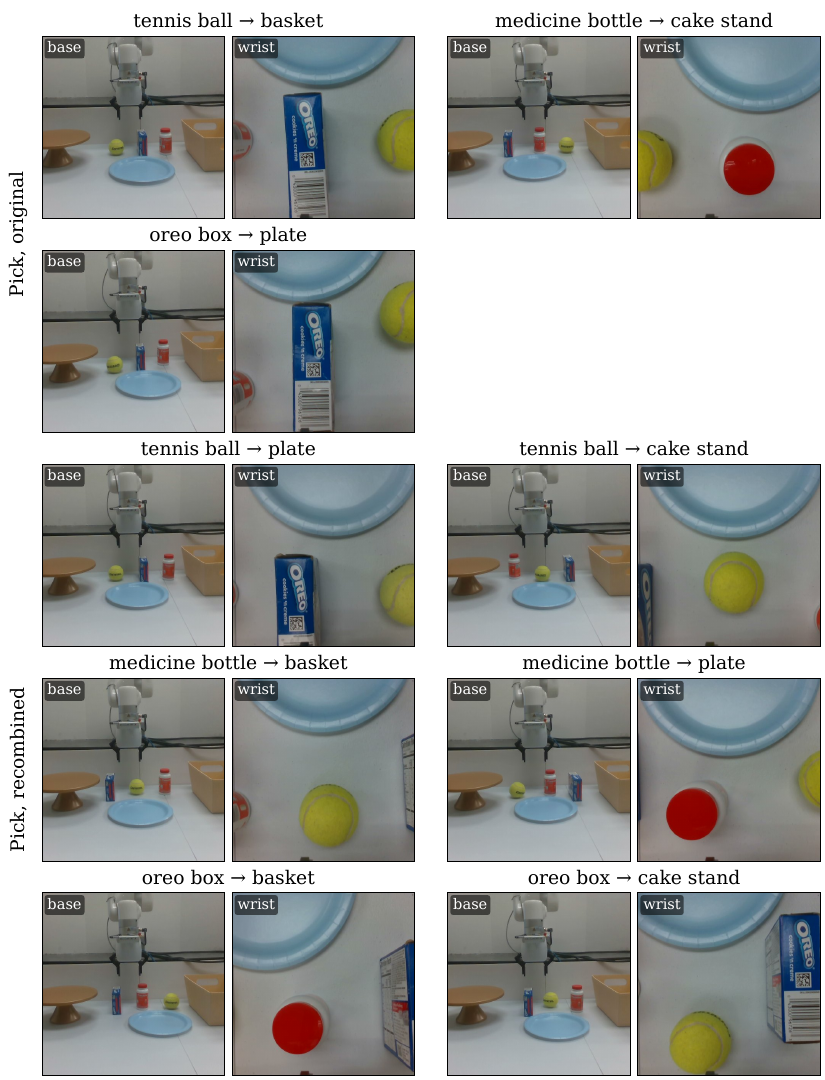}
\caption{One evaluation layout per pick-and-place cell with trained objects, as seen by the base camera (left) and the wrist camera (right) at the start of the episode. Rows group the original and the recombined pairings.}
\label{fig:real_cells_pick}
\end{figure}

\begin{figure}[htbp]
\centering
\includegraphics[width=\linewidth]{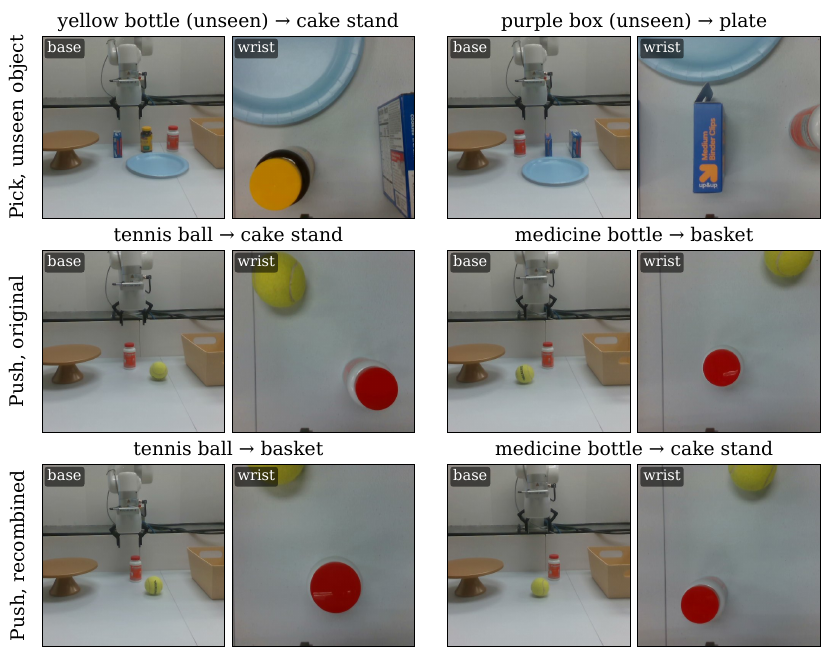}
\caption{One evaluation layout per unseen-object and push cell, same views as Fig.~\ref{fig:real_cells_pick}. In the unseen-object cells one training object is replaced by the unseen one.}
\label{fig:real_cells_unseen_push}
\end{figure}

A scripted demonstrator records 50 demonstrations for each of five tasks, 250 episodes and 59{,}167 frames at 10\,Hz. The pick-and-place tasks are tennis ball to basket, medicine bottle to cake stand and oreo box to plate, executed as hover, descend, close, lift, carry and place. The push tasks are tennis ball toward the cake stand and medicine bottle toward the basket, executed with a half-open gripper that descends behind the object and pushes it by 5\,cm along the line from the object to the destination. Events and poses are exported at 10\,Hz from the recorded observations and are the only input of the statistics of ReGuide, which are mined with the same code as in simulation.

\subsection{Evaluation Protocol}

The 15 cells of Table~\ref{tab:realcells} are evaluated under 10 layouts each with both methods, which gives 300 episodes. Both methods run the same client, and the bare policy is obtained by switching the wrapper off. Both methods are evaluated on the same paired initial layouts in each cell. In a cell with an unseen object, one randomly chosen training object is replaced by the unseen one, so three objects are on the table. An episode ends when the operator stops it and gives the verdict, at the step budget, or at a stop of the safety layer, and the last two count as failures. The budget is 600 policy steps after the 10 settle ticks in the pick-and-place cells (610 ticks in total) and 600 ticks including the settle ticks in the push cells (590 policy steps); tick counts reported for the robot include the settle ticks. Figs.~\ref{fig:real_cells_pick} and~\ref{fig:real_cells_unseen_push} show one layout per cell from both cameras.

\subsection{Grounding Module}

When an instruction is issued, GroundingDINO detects every known object and destination in the base image from a caption that contains their phrases. For each entity the most confident box is kept whose measured top height agrees with the declared height of the entity, and a box is owned by one entity only. During motion, the wrist image refines the position of an entity whenever it is in view, with the image paired to the tool pose at its capture time, and a fix is accepted only within 0.10\,m of the base estimate for objects and 0.25\,m for destinations. When a transport arrives, the target is measured once more and corrected by at most 60\,mm. The dimensions of the objects and destinations are declared from measurements, and those of the two unseen objects are measured before the evaluation. The 90th percentile of the horizontal error of the refined estimate over 14 recorded episodes is 2.6\,cm, and this value floors every horizontal tolerance that the demonstrations would set lower. There is no survey motion before an episode and no additional physical action for perception.

\subsection{Pushing}

\begin{figure}[htbp]
\centering
\includegraphics[width=\linewidth]{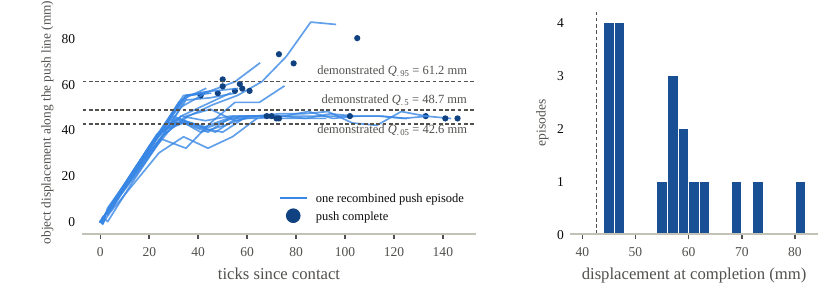}
\caption{Displacement of the object along the push line after contact in the 20 recombined push episodes (left) and its value when the push is declared complete (right). The dashed lines are the demonstrated quantiles. In 19 episodes the completion line is reached, and one episode is stopped by the operator at 45\,mm with the object already past the lower quantile.}
\label{fig:real_push}
\end{figure}

For pushing, the instructed object--destination line defines the interaction reference, with longitudinal, lateral and vertical axes. From the 100 push demonstrations, the pre-contact configuration lies 38.2\,mm behind the object along the line, 0.3\,mm lateral and 29.3\,mm below the object centre, with a tolerance of 10.6\,mm, a trigger radius of 2.4\,mm, a hover altitude of 279.7\,mm and a posture tolerance of 0.024\,rad. The demonstrated displacement has $Q_{.05}$, $Q_{.5}$ and $Q_{.95}$ of 42.6, 48.7 and 61.2\,mm, and a push is complete when the displacement of the object has exceeded the lower value for $n$ ticks. A commitment toward the habitual destination indicates a direction mismatch. After approach transport establishes the pre-contact configuration, carry-side transport advances along the instructed line by the median demonstrated displacement under the step bound. On original cells no transport is issued. During a push the wrapper keeps the tool on the demonstrated line and height. Fig.~\ref{fig:real_push} shows the displacement of the object after contact in the 20 recombined push episodes.

\subsection{Results}

\begin{figure}[htbp]
\centering
\includegraphics[width=\linewidth]{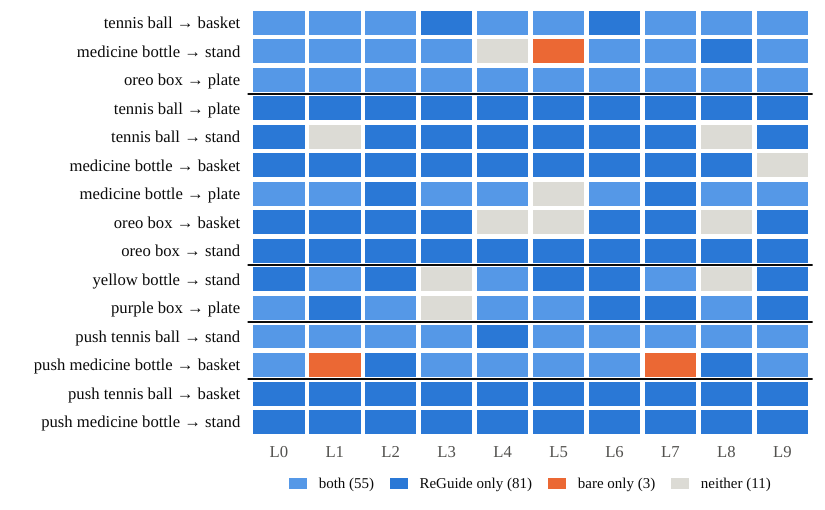}
\caption{Outcome of the 150 paired layouts of the real-robot study. Each square is one layout of one cell, coloured by whether both methods, only ReGuide, only the bare policy or neither succeeded.}
\label{fig:real_layouts}
\end{figure}

\begin{figure}[htbp]
\centering
\includegraphics[width=\linewidth]{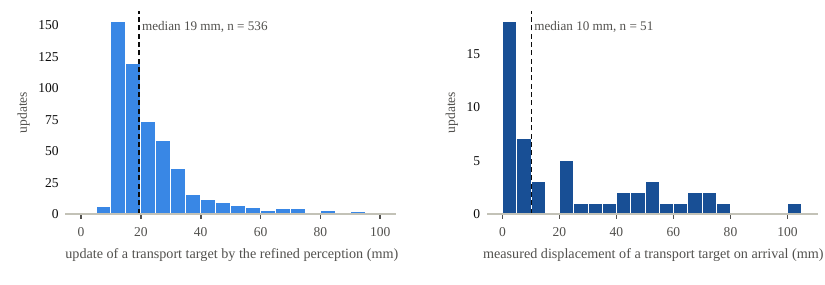}
\caption{Left, the update of a transport target by the refined perception, in motion and on arrival, 536 updates over the 110 pick-and-place episodes of ReGuide with a median of 19\,mm. Right, the displacement of the target measured by the arrival refinement, 51 measurements.}
\label{fig:real_perception}
\end{figure}

\begin{table}[htbp]
\centering
\caption{Real-robot successes out of 10 layouts per cell, and paired outcomes per block with the exact McNemar test.}
\label{tab:realcells}
\small
\fitwidth{%
\begin{tabular}{llcc}
\toprule
\textbf{Block} & \textbf{Object and destination} & \textbf{Bare $\pi_{0.5}$} & \textbf{ReGuide} \\
\midrule
Pick, original & tennis ball to basket & 8/10 & 10/10 \\
 & medicine bottle to stand & 8/10 & 8/10 \\
 & oreo box to plate & 10/10 & 10/10 \\
\multicolumn{2}{r}{\textit{paired layouts, both / bare only / ReGuide only / neither}} & \multicolumn{2}{c}{25 / 1 / 3 / 1, $p$ = 0.625} \\
\midrule
Pick, recombined & tennis ball to plate & 0/10 & 10/10 \\
 & tennis ball to stand & 0/10 & 8/10 \\
 & medicine bottle to basket & 0/10 & 9/10 \\
 & medicine bottle to plate & 7/10 & 9/10 \\
 & oreo box to basket & 0/10 & 7/10 \\
 & oreo box to stand & 0/10 & 10/10 \\
\multicolumn{2}{r}{\textit{paired layouts, both / bare only / ReGuide only / neither}} & \multicolumn{2}{c}{7 / 0 / 46 / 7, $p<0.001$} \\
\midrule
Pick, unseen object & yellow bottle to stand & 3/10 & 8/10 \\
 & purple box to plate & 5/10 & 9/10 \\
\multicolumn{2}{r}{\textit{paired layouts, both / bare only / ReGuide only / neither}} & \multicolumn{2}{c}{8 / 0 / 9 / 3, $p$ = 0.004} \\
\midrule
Push, original & tennis ball toward stand & 9/10 & 10/10 \\
 & medicine bottle toward basket & 8/10 & 8/10 \\
\multicolumn{2}{r}{\textit{paired layouts, both / bare only / ReGuide only / neither}} & \multicolumn{2}{c}{15 / 2 / 3 / 0, $p$ = 1.000} \\
\midrule
Push, recombined & tennis ball toward basket & 0/10 & 10/10 \\
 & medicine bottle toward stand & 0/10 & 10/10 \\
\multicolumn{2}{r}{\textit{paired layouts, both / bare only / ReGuide only / neither}} & \multicolumn{2}{c}{0 / 0 / 20 / 0, $p<0.001$} \\
\bottomrule
\end{tabular}%
}
\end{table}

\begin{figure}[htbp]
\centering
\includegraphics[width=0.8\linewidth]{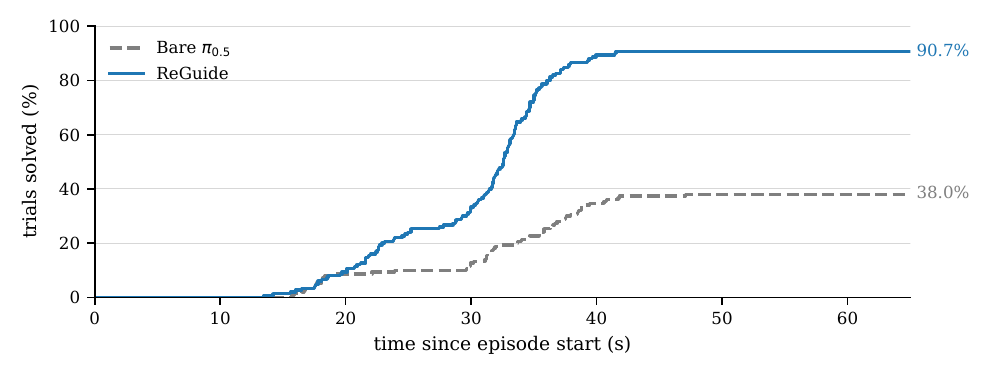}
\caption{Share of real-robot trials solved within a given time. One successful bare trial on an original cell has no recorded duration and is not drawn.}
\label{fig:realtime}
\end{figure}

Table~\ref{tab:realcells} gives the results per cell with the paired outcomes per block, and Fig.~\ref{fig:realtime} the share of trials solved within a given time. On the robot, guidance also supports execution on original tasks. The approach transport starts at a median of tick 67 on trained objects and at tick 10 on unseen objects, where first-decision guidance establishes the transferred configuration. On the original push cells the wrapper issues no transport. Fig.~\ref{fig:real_layouts} shows the paired outcome of every layout, and Fig.~\ref{fig:real_perception} the corrections of the transport targets.

\subsection{Correspondence between Simulation and Robot}

The robot instantiates referent monitoring and geometric guidance through its sensing and control interfaces. Entity poses come from the detector rather than simulator state, with measured perception error flooring the tolerances. Robot joint targets are converted by forward kinematics with unit gains, whereas simulation uses end-effector increments with fitted gains. Control runs at 10 rather than 20\,Hz, with event windows matched in seconds. Held and contact predicates use the gripper and tool position on the robot. Success is judged by the operator on the robot and by the goal predicate in simulation.

\section{Limitations}
\label{app:limitations}

ReGuide relies on the frozen policy's local interaction skills after hand-back, so failures in these skills remain a source of error. The failures that remain in Table~\ref{tab:residual} concentrate on such cells, for example the moka pot of B1, which GR00T N1.7 and OpenVLA-OFT fail in 48 of 50 trials while $\pi_{0.5}$ fails in one. ReGuide also relies on an external grounding module for the identity and the pose of the instructed referents. The main simulation comparison uses oracle grounding to isolate the wrapper from perception errors. Perception-based simulation (Appendix~\ref{app:perception_grounding}) and real-robot experiments additionally evaluate these errors. The demonstration statistics require the training demonstrations with end-effector and object poses, which not every released policy provides. Our evaluation covers pick-and-place and push tasks in which an instruction resolves to an object and a destination, and every composition cell changes one factor. Articulated or deformable manipulation, compositions that change several factors at once, and relational or ambiguous instructions are not evaluated. An unseen object inherits the statistics of the trained object with the closest shape, which does not cover objects that need a grasp absent from the demonstrations. Finally, the real-robot study uses one arm, one backbone and 150 paired trials.

\end{document}

%% file: math_commands.tex
\usepackage{amsmath,amsfonts,bm}

\def\eqref#1{equation~\ref{#1}}

\def\1{\bm{1}}

\DeclareMathAlphabet{\mathsfit}{\encodingdefault}{\sfdefault}{m}{sl}
\SetMathAlphabet{\mathsfit}{bold}{\encodingdefault}{\sfdefault}{bx}{n}

